\documentclass[10pt,twocolumn,letterpaper]{article}

\usepackage{times}
\usepackage{epsfig}
\usepackage{graphicx}
\usepackage{booktabs}
\usepackage{mathtools}

\usepackage{siunitx}
\usepackage{amsmath}
\usepackage{cases}
\usepackage{multirow}
\usepackage{amssymb}
\usepackage{dsfont}  %
\usepackage{arydshln}
\usepackage{iccv}              %

\usepackage{listings}

\lstdefinestyle{pythonstyle}{
    language=Python,
    basicstyle=\ttfamily\footnotesize,
    keywordstyle=\color{blue},
    commentstyle=\color{gray},
    stringstyle=\color{red},
    numberstyle=\tiny\color{gray},
    numbers=left,
    stepnumber=1,
    breaklines=true,
    breakatwhitespace=false,
    tabsize=4,
    frame=tb,
    captionpos=b
}

\definecolor{iccvblue}{rgb}{0.21,0.49,0.74}
\usepackage[pagebackref,breaklinks,colorlinks,allcolors=iccvblue]{hyperref}
\def\paperID{394} %

\def\confName{ICCV}
\def\confYear{2025}

\begin{document}

\title{VolumetricSMPL: A Neural Volumetric Body Model for\\Efficient Interactions, Contacts, and Collisions} 

\author{
Marko Mihajlovic$^{1}$, Siwei Zhang$^1$, Gen Li$^1$, Kaifeng Zhao$^{1}$, Lea M\"{u}ller$^{2}$, Siyu Tang$^1$\\
$^1$ETH Z\"{u}rich \  $^2$UC Berkeley \\[4pt]
  {\href[pdfnewwindow=true]{https://markomih.github.io/VolumetricSMPL}{\nolinkurl{markomih.github.io/VolumetricSMPL}}}\\[4pt]
}

\twocolumn[{%
    \renewcommand\twocolumn[1][]{#1}%
    \setlength{\tabcolsep}{0.0mm} %

    \maketitle
    \begin{center}
        \vspace{-2.5em}
        \newcommand{\teaserwidth}{\textwidth}
            \includegraphics[width=0.92\linewidth]{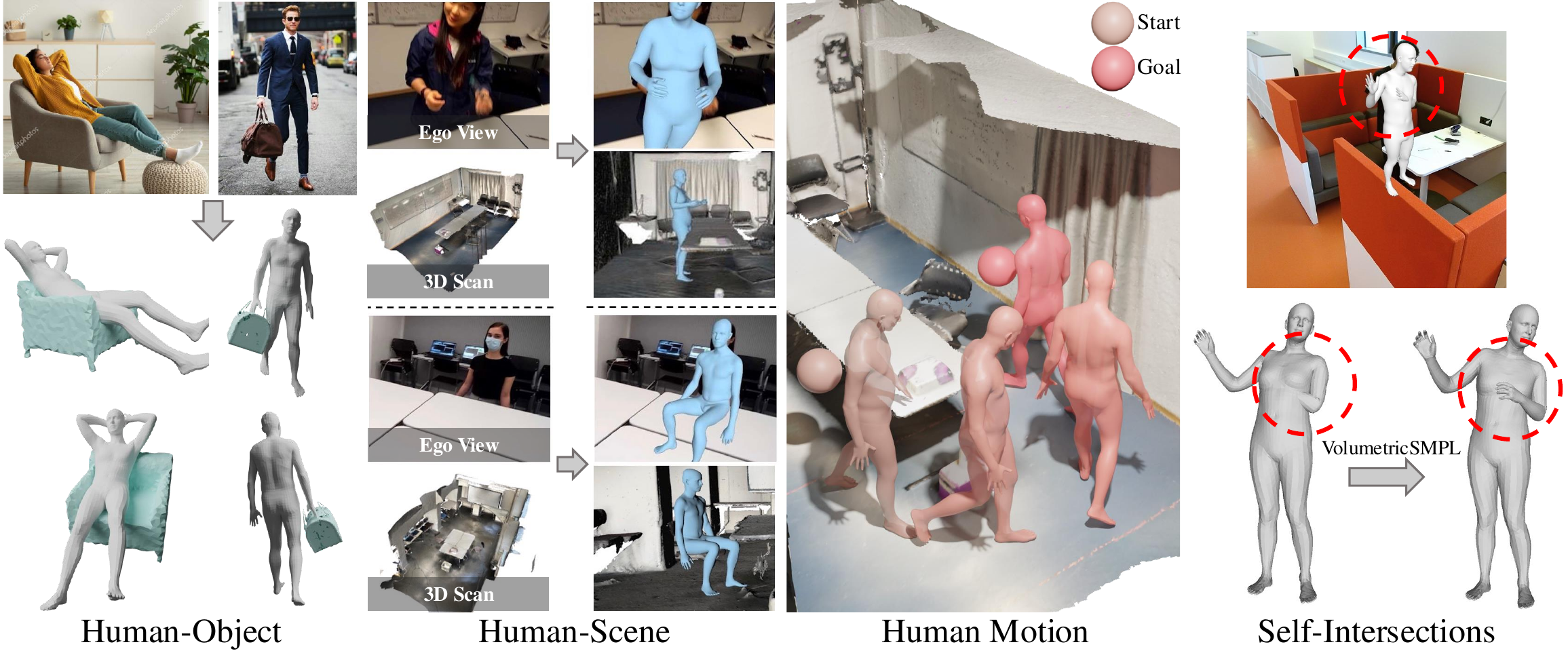}
      \vspace{-0.5em}
        \captionof{figure}{
We introduce VolumetricSMPL, a volumetric body model that represents human shape as a Signed Distance Field (SDF), enabling efficient, differentiable, and more accurate interactions across multiple downstream tasks:
(1) [\cref{sec:hoi}]: Human-object interaction reconstruction from in-the-wild images, achieving 500× speedup over prior work~\cite{zhang2020phosa, wang2022reconstruction}. 
(2) [\cref{subsec:ego_hmr}]: 3× faster and more accurate human mesh recovery in 3D scenes from egocentric images compared to~\cite{zhang2023probabilistic}.
(3) [\cref{subsec:human_motion}]: 7× faster scene-constrained human motion synthesis~\cite{Zhao:DART:2024} with 20× lower GPU memory, leading to more realistic and collision-free motion compared to imposing scene-constrains via prior work \cite{mihajlovic2022coap}. 
(4) [\cref{subsec:exp_selfinter}]: Efficient volumetric constraints for resolving mesh self-intersections, achieving 3× speedup, 3× lower GPU memory, and improved intersection handling over~\cite{mihajlovic2022coap}.
VolumetricSMPL provides a powerful solution for learning-based human interaction modeling with improved efficiency, scalability, and accuracy. 
} 

    \label{fig:teaser}
    \end{center}%
}]

\begin{abstract}
Parametric human body models play a crucial role in computer graphics and vision, enabling applications ranging from human motion analysis to understanding human-environment interactions.
Traditionally, these models use surface meshes, which pose challenges in efficiently handling interactions with other geometric entities, such as objects and scenes, typically represented as meshes or point clouds. 
To address this limitation, recent research has explored volumetric neural implicit body models. 
However, existing works are either insufficiently robust for complex human articulations or impose high computational and memory costs, limiting their widespread use.
To this end, we introduce VolumetricSMPL, a neural volumetric body model that leverages Neural Blend Weights (NBW) to generate compact, yet efficient MLP decoders. Unlike prior approaches that rely on large MLPs, NBW dynamically blends a small set of learned weight matrices using predicted shape- and pose-dependent coefficients, significantly improving computational efficiency while preserving expressiveness.
VolumetricSMPL outperforms prior volumetric occupancy model COAP with 10× faster inference, 6× lower GPU memory usage, enhanced accuracy, and a Signed Distance Function (SDF) for efficient and differentiable contact modeling.
We demonstrate VolumetricSMPL’s strengths across four challenging tasks: (1) reconstructing human-object interactions from in-the-wild images, 
(2) recovering human meshes in 3D scenes from egocentric views, 
(3) scene-constrained motion synthesis, and 
(4) resolving self-intersections. 
Our results highlight its broad applicability and significant performance and efficiency gains. 
\end{abstract}

\section{Introduction} \label{sec:intro}
Learned parametric body models \cite{SMPL:2015,pavlakos2019expressive,osman2020star,xu2020ghum} are essential tools in computer graphics and vision, supporting diverse applications such as avatar creation \cite{MetaAvatar:NeurIPS:2021,xu2021hnerf,kolotouros2023dreamhuman,yuan2023gavatar}, motion analysis \cite{sminchisescu2002human, LEMO:Zhang:ICCV:2021, rempe2021humor, MOJO:CVPR:2021, yuan2023physdiff}, human pose and shape reconstruction \cite{tian2023recovering}, and even medical treatments \cite{hesse2019learning,guo2022smpl}. 
These models are typically constructed by fitting parametric meshes to large-scale body scan datasets, providing a compact yet expressive representation of body shape and pose-dependent deformations.

Despite their efficiency in rendering and memory usage, mesh-based deformable body models face challenges in efficient intersection testing and differentiable contact modeling, particularly when interacting with objects, environments, and other humans. 
Existing methods address this limitation with application-specific strategies. 
Some preprocess 3D scenes into volumetric representations~\cite{PROX:2019,LEMO:Zhang:ICCV:2021,zhang2020generating,PLACE:3DV:2020}, which is computationally expensive and error-prone. Others constrain interactions to synthetic environments where analytic distance fields can be computed~\cite{hassan2021populating,Zhao:DART:2024}, limiting real-world applicability. 
Techniques modeling human-human interactions \cite{muller2024generative,fieraru2021remips} often substantially downsample meshes to enable traditional graphics-based methods, such as winding numbers~\cite{jacobson2013robust}, but these are non-differentiable and difficult to generalize to other applications that require differentiable collision loss terms~\cite{zhang2023probabilistic, li2024interdance, wang2022reconstruction, zhang2020phosa}. Overall, these solutions tend to be computationally costly, application-specific, or non-differentiable, hindering their broader adoption. 

To overcome these challenges, volumetric human body models~\cite{LEAP:CVPR:21,alldieck2021imghum,mihajlovic2022coap} have emerged as a unified solution that facilitates more natural and efficient interactions between the human body and point clouds or meshes. 
LEAP~\cite{LEAP:CVPR:21} represents the body as a volumetric function in canonical space, using a deformation network to map between observed and canonical coordinates. 
imGHUM~\cite{alldieck2021imghum} models the body directly in observation space using sub-part learnable neural implicit fields, though such a flexible model requires large collections of human scans for training. 
COAP~\cite{mihajlovic2022coap} introduces compositional neural fields in the canonical space of individual body parts and explicitly models deformations via skeletal transformations, improving pose generalization. 
Due to its strong performance, COAP has been widely adopted in applications such as human-scene interaction, human-object interaction, and digital human synthesis~\cite{li2024lodge_plus_plus,armani2024ultra,xia2024envposer,dai2025interfusion,kim2025beyond,li2024interdance,li2024egogen,zhang2023probabilistic}.

However, COAP’s large MLP decoder creates a computational bottleneck, restricting its widespread usability in learning- and optimization-based applications. 
Querying COAP with a batch size of 5 and 15k query points fully saturates a 24 GB GPU, making it impractical for integration with additional neural modules. Moreover, COAP represents the body as a step occupancy function, which leads to ill-defined gradients for points far from the iso-surface, restricting its effectiveness in tasks requiring smooth distance approximation such as scene-constrained motion synthesis demonstrated in \cref{subsec:human_motion}. 

To overcome these limitations, we introduce VolumetricSMPL, a novel volumetric body model that not only offers improved accuracy over COAP but also significantly enhances computational efficiency.
VolumetricSMPL achieves around \emph{10× faster} inference, requires \emph{6× less GPU memory}, and models the human body as a Signed Distance Field (SDF) instead of an occupancy function, ensuring smooth gradients for downstream tasks. 

At the core of VolumetricSMPL is the Neural Blend Weights (NBW) Generator, a neural network that dynamically predicts compact MLP weights for an SDF decoder. 
Unlike prior volumetric body models~\cite{mihajlovic2022coap,LEAP:CVPR:21,alldieck2021imghum}, which rely on large MLP decoders with 256–512 neurons, NBW generates lightweight 64-neuron MLPs by blending compact learned weight matrices with predicted shape- and pose-dependent coefficients. 
This design preserves expressiveness while drastically reducing computational cost, leading to faster inference, improved accuracy, and lower memory usage. 

VolumetricSMPL builds upon COAP’s compositional encoder, partitioning the body into kinematic chain-based parts \cite{mihajlovic2022coap}, each represented in its own local canonical space and encoded using a lightweight PointNet MLP \cite{qi2017pointnet}.
The NBW module then generates specialized network weights for each local SDF decoder, effectively creating part-specific networks, akin to the mixture of experts paradigm~\cite{yuksel2012mixtureofexperts}.
At inference time, each local decoder predicts a signed distance for a given query point, and the results are aggregated using a minimum operation, ensuring a smooth, globally consistent SDF representation.
To further boost efficiency, we introduce an analytic SDF for regions far from the body, significantly accelerating queries in non-critical areas while maintaining accuracy near the surface.

We extensively evaluate VolumetricSMPL and demonstrate its superior performance across multiple metrics compared to existing models. Furthermore, we showcase its effectiveness in four key downstream applications:
\textit{1)} reconstructing human-object interactions from in-the-wild images~\cite{wang2022reconstruction},
\textit{2)} recovering human meshes in 3D scenes from egocentric views~\cite{zhang2023probabilistic},
\textit{3)} scene-constrained human motion synthesis, and
\textit{4)} resolving self-intersections in human body models~\cite{mihajlovic2022coap}. 
These applications highlight significant improvements in speed, accuracy, and convergence, enabled by our efficient SDF representation (see \cref{fig:teaser}).

In summary, our contributions are:
\begin{itemize}[leftmargin=0.6cm,itemsep=0pt, topsep=0pt, parsep=0pt]
    \item A novel parametric Signed Distance Field body model, achieving around \emph{10× faster} inference, \emph{6× lower GPU memory usage}, and \emph{higher accuracy} compared to COAP~\cite{mihajlovic2022coap}.  
    \item The introduction of Neural Blend Weights Generator that enables efficient SDF learning with tiny MLPs. 
    \item The deployment of VolumetricSMPL, leading to substantial computational efficiency and accuracy gains across downstream applications, including human-object interaction reconstruction, human mesh recovery, scene-constrained human motion synthesis, and self-intersection resolution, demonstrating its practicality and versatility. See \cref{fig:teaser} as an overview. 
\end{itemize}
We provide a lightweight and easy-to-use add-on module for the widely used SMPL-based body models \cite{SMPL:2015, pavlakos2019expressive}, enabling seamless integration of VolumetricSMPL with a single line of code, without compromising the generalization of our method. The code and models will be released under the MIT license.

\section{Related Work}

\textbf{Data-Driven Body Models} are important components in computer vision and graphics \cite{cheng2018parametric, SMPL:2015, pavlakos2019expressive, pishchulin2017building, xu2020ghum}. Trained on large-scale human scan datasets, these models generate human shapes using low-dimensional representations, such as shape coefficients and pose parameters. 
Many parametric models employ linear shape spaces \cite{SMPL:2015, pavlakos2019expressive, MANO:SIGGRAPHASIA:2017,FLAME:SiggraphAsia2017}, ensuring compatibility with graphics pipelines while maintaining high-quality shape representation. 
We develop our volumetric representation based on the SMPL body model \cite{SMPL:2015, pavlakos2019expressive}, inheriting its generative capabilities. This choice ensures seamless integration with prior methods operating within the SMPL parameter space \cite{tian2023recovering}, facilitating compatibility with existing shape estimation and reconstruction pipelines. 

\textbf{Modeling Human Interactions} faces challenges for mesh body models~\cite{PROX:2019, bhatnagar22behave, Zhang:ECCV:2022, zhang2020generating, Zhao:ICCV:2023, Zhao:ECCV:2022, zhang2022couch}. 
Many approaches rely on application-specific solutions. 
To model human-scene interactions, several methods~\cite{jiang2020coherent, khirodkar2022occluded, PROX:2019, zhang2020generating, li2024egogen} employ additional preprocessing steps to convert 3D scans into volumetric representations, enabling efficient collision detection and resolution. However, these techniques are often computationally expensive, memory-intensive, and prone to inaccuracies. To mitigate these limitations, some works~\cite{yuan2023physdiff, zhang2022wanderings, karunratanakul2023optimizing, luo2021dynamics, yuan2021simpoe} constrain interactions to synthetic environments or simplified geometric primitives, though this often comes at the cost of reduced realism. Similarly, techniques for modeling human-human interactions \cite{muller2024generative,fieraru2021remips} often aggressively downsample meshes to make traditional graphics-based methods, such as winding numbers~\cite{jacobson2013robust}, computationally feasible. However, these methods remain non-differentiable and challenging to generalize to other applications that require differentiable collision loss terms~\cite{zhang2023probabilistic, li2024interdance, wang2022reconstruction, zhang2020phosa}. 

\textbf{Volumetric Body Models} address the limitations of mesh models in human interaction tasks. 
While classical tetrahedral meshes~\cite{hang2015tetgen} are used for collision detection, they lack direct support for animatable body models, as skinning is typically defined only on surface points and are rather used to model rigid body parts \cite{li2022nimble,kim2017vsmpl}. 
A more flexible alternative is neural volumetric body models~\cite{varol2018bodynet, nasa, palafox2021npm, palafox2022spams, LEAP:CVPR:21, mihajlovic2022coap, alldieck2021imghum}. 
BodyNet~\cite{varol2018bodynet} regresses a volumetric body directly from images, while other methods~\cite{nasa, palafox2021npm, palafox2022spams} require subject-specific optimization, making them unsuitable for feed-forward inference modeling of human bodies akin to SMPL. 
More closely related to our work, LEAP~\cite{LEAP:CVPR:21}, COAP~\cite{mihajlovic2022coap}, and imGHUM~\cite{alldieck2021imghum} enable feed-forward volumetric body modeling from low-dimensional shape and pose parameters. These models integrate seamlessly into perception applications~\cite{li2024lodge_plus_plus,armani2024ultra,xia2024envposer,dai2025interfusion,kim2025beyond}, facilitating direct interactions with scene and object representations.

\textbf{Feed-Forward Volumetric Body Models} \cite{LEAP:CVPR:21,mihajlovic2022coap,alldieck2021imghum} enable efficient human body modeling using low-dimensional shape and pose parameters while facilitating differentiable interactions with point clouds and meshes. Among them, imGHUM~\cite{alldieck2021imghum} learns expressive implicit neural fields directly in observation space, providing flexibility but requiring large-scale human scan datasets for training, which limits its adaptability to new shape spaces. 
In contrast, LEAP~\cite{LEAP:CVPR:21} and its more robust successor, COAP~\cite{mihajlovic2022coap}, restrict the learning space to the shape space of parametric mesh models such as SMPL~\cite{SMPL:2015} and SMPL-X~\cite{pavlakos2019expressive}. This design choice allows for seamless integration into downstream tasks that rely on these parametric models and require efficient body-environment interactions. Notably, COAP has already been widely adopted in applications such as human-scene and human-object interaction modeling~\cite{li2024lodge_plus_plus,armani2024ultra,xia2024envposer,dai2025interfusion,kim2025beyond,li2024interdance,li2024egogen,zhang2023probabilistic}.

\begin{figure*}[th!]
    \begin{center}
        \includegraphics[width=\textwidth]{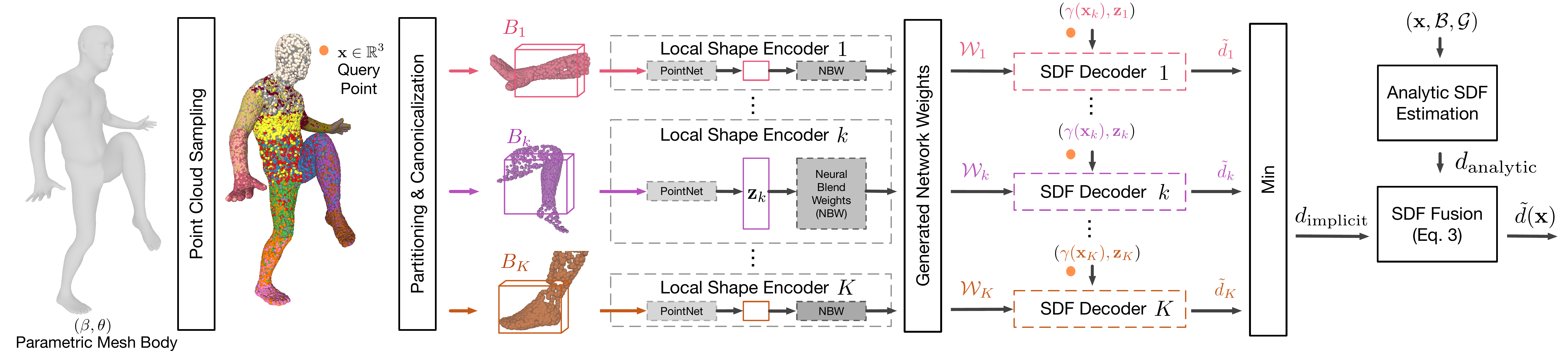}
    \end{center}
    \vspace{-1.0em}
    \caption{\textbf{Overview.} 
VolumetricSMPL models the human body volume as a Signed Distance Field (SDF) based on a parametric mesh body model~\cite{SMPL:2015, pavlakos2019expressive}, controlled via shape and pose parameters \( (\beta, \theta) \).  
First, the human mesh is sampled into a point cloud, where each point is assigned to one of \( K \) body parts (colored regions). Each part is then canonicalized based on its respective bone transformation \( G_k \) and over-approximated by a bounding box \( B_k \) following~\cite{mihajlovic2022coap}.  
Next, each body part is encoded into a compact feature vector \( \mathbf{z}_k \) via a PointNet MLP \cite{qi2017pointnet}, which is further processed by the Neural Blend Weights (NBW) generator (\cref{subsec:NBW}) that predicts weights \( \mathcal{W}_k \) for the corresponding local SDF decoder. 
For a given query point \( \mathbf{x} \in \mathbb{R}^3 \), we first transform it into its canonicalized form \( \mathbf{x}_k \) (Eq.~\ref{eq:x_k}) and optionally apply positional encoding \( \gamma(\cdot) \)~\cite{mildenhall2021nerf}. The SDF prediction \( \tilde{d}_k \) is then obtained by querying each local decoder, which takes as input both \( \mathbf{x}_k \) and the corresponding shape feature \( \mathbf{z}_k \). The local predictions are then aggregated using a min operation, yielding \( d_{\text{implicit}} \). If the query point is far from the body, outside of all boxes $\mathcal{B} = [B_k]_{k=1}^K$ placed via $\mathcal{G} = [G_k]_{k=1}^K$, an analytic SDF estimate \( d_{\text{analytic}} \) is used for efficiency. The final signed distance prediction \( \tilde{d}(\mathbf{x}) \) is computed by fusing the implicit and analytic estimates. 
Gray blocks indicate trainable neural modules. 
}    
    \label{fig:overview}
\end{figure*}

However, learning- and optimization-based applications require frequent queries to the volumetric body representation to model collisions, making computational efficiency a critical concern. We analyze existing volumetric body models and identify the primary bottleneck as the computational cost of large MLP decoders. For instance, both LEAP and COAP employ 256-neuron MLP decoders, while imGHUM relies on even larger 512-neuron architectures. 

\textbf{Generating Network Weights} from a conditioning vector has been explored via direct weight regression using HyperNetworks~\cite{Ha2017HyperNetworks} and weight modulation approaches such as FiLM~\cite{chan2021pi}. HyperNetworks have also been applied for subject-specific avatar modeling~\cite{MetaAvatar:NeurIPS:2021}. However, we find these approaches unsuitable for feed-forward modeling of compositional volumetric bodies, as each body part is only weakly supervised, leading to unstable training. 

Instead, we build on insights from ResFields \cite{ResFields}, which compose learned network weights blended through optimizable coefficients for fitting temporal signals. Here, we repurpose this design and regress pose- and shape-dependent blending coefficients, introducing our Neural Blend Weights (NBW) Generator to predict compact MLPs for efficient feed-forward volumetric body modeling.

\section{VolumetricSMPL}

VolumetricSMPL is a volumetric body model \( f_\Theta \), parameterized by learned parameters \( \Theta \) and conditioned on human shape \( \beta \in \mathbb{R}^{|\beta|} \) and pose \( \theta \in \mathbb{R}^{|\theta|} \), following the structure of widely used parametric human models~\cite{SMPL:2015, xu2020ghum, pavlakos2019expressive}. 
We denote \( K \) body parts using rigid bone transformation matrices \( \mathcal{G} = [G_k \in \mathbb{R}^{4 \times 4} ]_{k=1}^K \), which define the location and orientation of each body part. 
Formally, the volumetric body model is defined as a level-set function that maps a query point \( \mathbf{x} \in \mathbb{R}^3 \) to a signed distance value:
\begin{equation}\label{eq:f_theta}
    \tilde{d}(\mathbf{x}) = f_\Theta\left(\mathbf{x} \mid \beta,\theta\right) \in \mathbb{R}\,,
\end{equation}
where \( \tilde{d}(\mathbf{x}) \) represents the distance to the closest surface, with its sign indicating whether the point lies inside or outside the human body. An overview is provided in \cref{fig:overview}.

\subsection{Model Representation and Encoding} 
VolumetricSMPL extends a parametric mesh body model (\eg, SMPL~\cite{SMPL:2015, pavlakos2019expressive}) by representing human shape as a Signed Distance Field (SDF), controlled by its shape and pose parameters $(\beta, \theta)$. 

First, the human mesh is converted into a point cloud by uniformly sampling points proportional to mesh face areas. 
Each point is then assigned to one of $K$ body parts using a kinematic chain partitioning scheme~\cite{mihajlovic2022coap}. 
Each body part is then transformed into a local canonical space using its respective bone transformation \( G_k \) and over-approximated by a bounding box \( B_k \), constructed by extending the extremities of the part with a \( 12.5\% \) padding. 
To efficiently encode shape information, the point cloud of each body part is processed using a compact shared PointNet encoder~\cite{qi2017pointnet}. 
This produces a local feature vector $\mathbf{z}_k$, which captures shape and deformation information. This vector then conditions the Neural Blend Weights (NBW) module (\cref{subsec:NBW}), which generates adaptive network weights  $\mathcal{W}_k$ for the $k$th SDF decoder, enabling an efficient and flexible volumetric representation. 

\subsection{Efficient SDF Querying} \label{subsec:SDF_Querying}
To evaluate the SDF efficiently, we transform the query point $\mathbf{x}$ into the canonical space of each body part using a rigid transformation matrix that preserves signed distances:
\begin{equation} \label{eq:x_k} 
    \mathbf{x}_k = (G_k^{-1} \mathbf{x}^h)_{1:3}, \quad \text{ where } \quad \mathbf{x}^h=[\mathbf{x}, 1] \in \mathbb{R}^4.
\end{equation} 
Each local SDF decoder, using the generated weights \( \mathcal{W}_k \) from the NBW module and conditioned on \( \mathbf{z}_k \), predicts a signed distance \( \tilde{d}_k \) for the transformed query point. The final neural implicit SDF estimate \( d_{\text{implicit}} \) is obtained by taking the minimum over all local predictions. 

To improve computational efficiency, we introduce an analytic SDF approximation \( d_{\text{analytic}} \) for query points far from the human body. Specifically, if \( \mathbf{x} \) lies outside all bounding boxes $\mathcal{B}= [B_k]_{k=1}^K$, we approximate its signed distance using a bounding box-based SDF:
\begin{equation} \label{eq:d_union}
\tilde{d}(\mathbf{x}) = \begin{cases} 
    d_\text{analytic}\left(\mathbf{x} \mid \mathcal{B}, \mathcal{G} \right), & \text{if } \mathbf{x} \notin B_k \text{ for all } k, \\
    d_\text{implicit} \left(\mathbf{x} \mid \beta, \theta \right), & \text{otherwise}.
\end{cases}
\end{equation}
The analytic SDF computes the Euclidean distance to the closest bounding box surface, enabling a fast, memory-efficient approximation for distant points. This hybrid approach offers two key advantages:  
\emph{(i) Improved scalability and accuracy}, as relying solely on neural models is computationally expensive and prone to inaccuracies in distant regions; and  
\emph{(ii) Faster querying}, as the analytic SDF efficiently approximates distances, it reduces neural computation in non-critical areas. 

\subsection{Neural Blend Weights Generator} \label{subsec:NBW}
\begin{figure}[t!]
    \begin{center}
        \includegraphics[width=\linewidth]{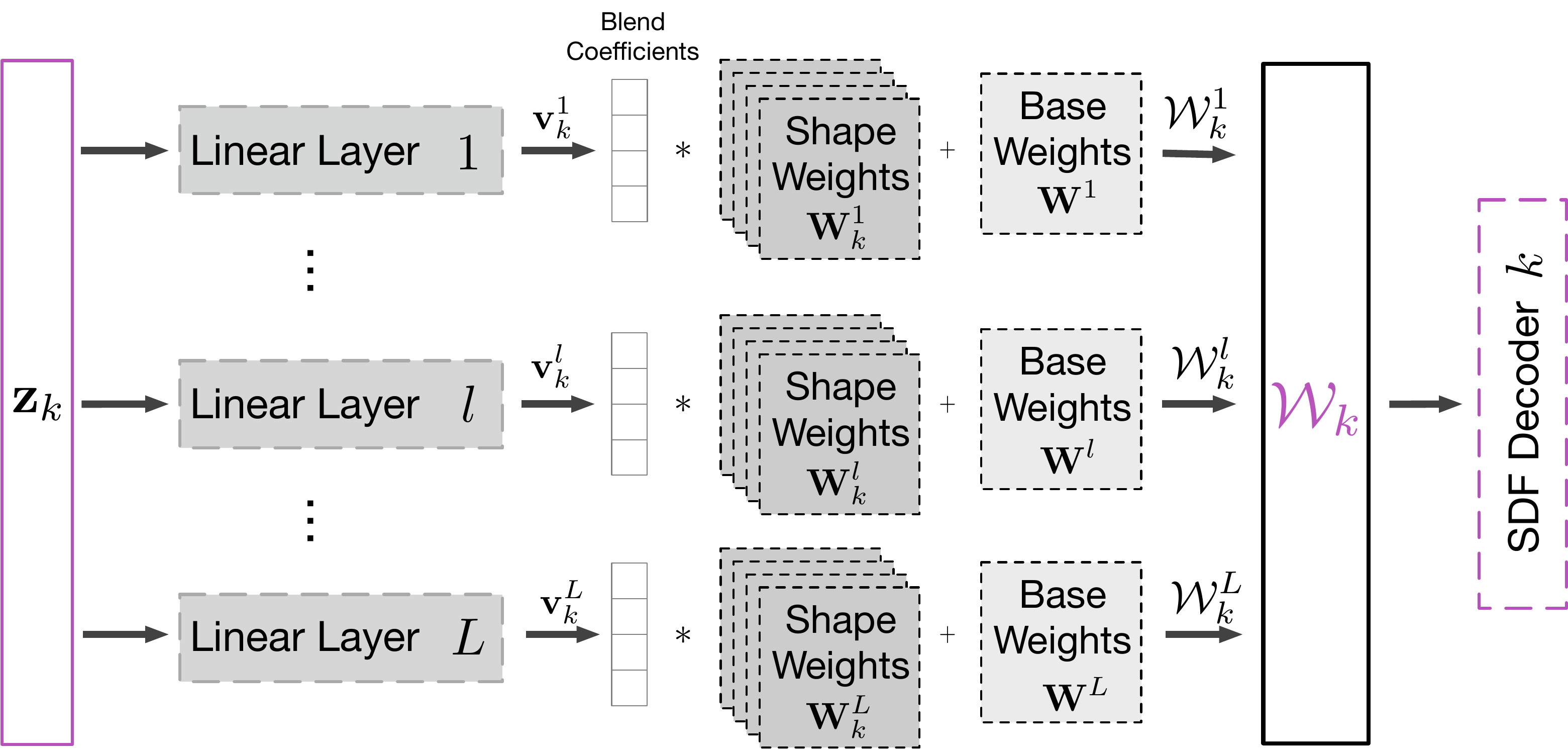}
    \end{center}
    \caption{\textbf{Neural Blend Weights (NBW) Generator} takes a local latent code \( \mathbf{z}_k \) as input and outputs layer-specific weights \( \mathcal{W}_k \) for the \( k \)th SDF decoder.  
First, \( \mathbf{z}_k \) is transformed into a set of blending coefficients \( \{\mathbf{v}_k^l\}_{l=1}^L \) using \( L \) independent learnable linear layers. These coefficients modulate the shape weights \( \mathbf{W}_{k}^{l} \), which are then added to the base network weights \( \mathbf{W}^l \).  
The resulting neural blend weights \( \mathcal{W}_k = \{\mathcal{W}_k^l\}_{l=1}^L \) are then used in the MLP-based SDF decoder. For brevity, bias terms are omitted. Gray blocks indicate trainable modules.  
    }
    \label{fig:NBW}
\end{figure}

Following the success of neural fields~\cite{xie2021neuralfields} in representing continuous SDF functions via MLP architectures, we implement an MLP-based SDF decoder. This decoder maps a query point \( \mathbf{x}_k \) (optionally positionally encoded as \( \gamma(\mathbf{x}_k) \)~\cite{mildenhall2021nerf}) and a conditioning latent code \( \mathbf{z}_k \) to an SDF value. Formally, our $k$th \( L \)-layer SDF decoder is defined as: 
\begin{equation} \label{eq:mlp}
    \tilde d_k = \sigma_L \left(\mathcal{W}_k^L (\phi_{L-1} \circ \cdots \circ \phi_1)(\gamma(\mathbf{x}_k),\mathbf{z}_k) + \mathbf{b}_L\right),  
\end{equation}
where each layer \( \phi_l: \mathbb{R}^{M_l} \mapsto \mathbb{R}^{N_l} \) applies affine transformation over the the layer input \( \mathbf{x}_l^{\prime} \in \mathbb{R}^{M_l} \) followed by a non-linear activation function \( \sigma_l \):  
\begin{equation} \label{eq:mlp_layer}
    \phi_l(\mathbf{x}_l^{\prime}) = \sigma_l\left(\mathcal{W}_k^l\mathbf{x}_l^{\prime}+\mathbf{b}_l\right).
\end{equation}
Here, \( \mathcal{W}_k^l \) and \( \mathbf{b}_l \in \mathbb{R}^{N_l} \) are the weights and biases specific to the $l$th layer. 

\textbf{The Challenge: Balancing Efficiency and Expressiveness.}
A naive approach would treat \( \mathcal{W}_k^l \in \mathbb{R}^{N_l \times M_l} \) as fully trainable, as done in prior volumetric models~\cite{LEAP:CVPR:21, mihajlovic2022coap, alldieck2021imghum}. However, to capture complex human shapes and poses, these models require wide MLPs (256–512 neurons per layer), leading to:
1) \emph{High computational cost} – large networks slow down inference.
2) \emph{Memory inefficiency} - wide weight matrices consume excessive GPU memory. 

To address this, we propose the Neural Blend Weights generator (NBW) that predicts efficient neural network weights for narrow MLPs that retain expressiveness while being compact. 
An overview of NBW is illustrated in \cref{fig:NBW}, showing how \( \mathbf{z}_k \) is transformed into MLP weights.

\textbf{How NBW Works.}
For each layer \( l \) of the MLP, we define the weight matrix \( \mathcal{W}_k^l \) as a linear combination of $R$ learnable base weight matrices $\mathbf{W}_k^l[r] \in \mathbb{R}^{N_l \times M_l}$ (where $r=1,\ldots,R$):
\begin{equation}
    \mathcal{W}_k^l = \mathbf{W}^l + \sum\nolimits_{r=1}^{R} \mathbf{v}_k^l[r] \mathbf{W}_k^l[r] .
\end{equation}
Here,
\( \mathbf{W}^l \in \mathbb{R}^{N_l \times M_l} \) is the base MLP weight matrix.
\( \mathbf{W}_k^l[r] \) are $R$ learned shape weight matrices specific to the $k$th body part and layer $l$.
\( \mathbf{v}_k^l \in \mathbb{R}^{R} \) are blending coefficients that determine how much each base matrix contributes to the final weights.

These blending coefficients $\mathbf{v}_k^l$ are not fixed; instead, they are computed dynamically by passing the latent code $\mathbf{z}_k$ -- which encodes the shape and pose of the $k$th body part -- through an independent linear layer. 
This ensures that the MLP weights adapt to the specific body configuration. 

This approach draws inspiration from the ResField architecture~\cite{ResFields}, originally designed for temporal signal fitting, but here repurposed for a feed-forward inference of network weights for volumetric body modeling. 

\textbf{Advantages of NBW}:
1) \emph{Compact architecture} – NBW allows small MLPs ($64 \times 64$ weight matrices), drastically reducing memory usage and speeding up inference.
2) \emph{High expressiveness} – Dynamic blending of weights enables accurate modeling of complex SDFs, rivaling larger MLPs.
3) \emph{Scalability} – Lower GPU memory consumption enables practical use in real-world applications. 

\subsection{Training} \label{subsec:training}
We follow the training protocol established in \cite{LEAP:CVPR:21, mihajlovic2022coap}, using SMPL \cite{SMPL:2015} body model meshes from the AMASS dataset \cite{AMASS:ICCV:2019}. For training, we sample every 5th body model from the MoVi dataset \cite{ghorbani2020movi} and every 10th model from DFaust \cite{dfaust:CVPR:2017}. 
For validation, we use PosePrior~\cite{PosePrior_Akhter:CVPR:2015}, a dataset featuring challenging body poses, where every 300th model is selected, resulting in a total of 316 bodies. 

At each training step, we sample query points  $\mathcal{D}$ following the approach in \cite{mihajlovic2022coap}. Half of the samples (256 per body part) are drawn uniformly inside the bounding boxes $\mathcal{B}$, while the other half is sampled around the mesh surface using Gaussian noise $\sim \mathcal{N}(0, 0.1)$. The ground truth SDF value for each query point $d(\mathbf{x})$ is estimated by computing its distance to the surface mesh, with the sign robustly determined via inside-outside checks as described in \cite{mihajlovic2022coap, nasa, LEAP:CVPR:21}. 

The model is trained by supervising both the sign and absolute value of the SDF. The loss function is defined as: 
\begin{equation} \label{eq:l2_training}
    \mathcal{L} = 
        \sum_{\textbf{x} \in \mathcal{D}} l_2\left(sgn(\tilde{d}(\textbf{x})), sgn(d(\textbf{x}))\right) + l_2\left(|\tilde{d}(\textbf{x})| , |d(\textbf{x})|\right) \,.
\end{equation}
We train the network using a batch size of 8, optimizing the weights by minimizing $\mathcal{L}$ (Eq.~\ref{eq:l2_training}) with the Adam optimizer~\cite{kingma2014adam}. The learning rate is initialized at $10^{-4}$ and gradually annealed to $10^{-5}$. Training is performed for 15 epochs, totaling 450k iterations, and takes approximately 20 hours on a single 24GB RTX 3090 GPU. 

\section{Experiments} 
We evaluate VolumetricSMPL by benchmarking it against recent volumetric body models (\cref{subsec:exp_bodies}) and demonstrating its effectiveness across multiple downstream applications (\cref{subsec:exp_applications}) visualized in \cref{fig:teaser}.

\begin{table}[tb]
  \caption{
    \textbf{Comparisons with Volumetric Bodies.} 
    Compared to COAP~\cite{mihajlovic2022coap} on unseen subjects from the challenging PosePrior~\cite{PosePrior_Akhter:CVPR:2015} dataset. 
    Our body model is more accurate while providing SDF representation, being around 10× times faster, and using 6× times less GPU memory. 
    The resources are measured on a total of 60k query points uniformly sampled around the body and sampled around the surface; 
    \textit{surf.} and \textit{unif.} IoU refers to points sampled around the surface and uniformly at random respectively.
  }
  \label{tab:bodies}
  \centering
  \setlength{\tabcolsep}{5.0pt} %

\resizebox{\linewidth}{!}{
\begin{tabular}{@{}lcc|ccc|cc@{}}
\toprule
& \multicolumn{2}{c}{$\text{Resources}$} 
& \multicolumn{3}{c}{$\text{IoU [\%]}~\uparrow$} 
& \multicolumn{2}{c}{$\text{MSE}~\downarrow$} \\
& $\text{t [ms]}~\downarrow$ & $\text{GPU}~\downarrow$ & mean & \textit{surf.} & \textit{unif.} & SDF & $\lvert \text{SDF} \rvert$ \\
\midrule
LEAP~\cite{LEAP:CVPR:21}            & \phantom{0}79                 & \phantom{0}7.7 G & 75.98 & 69.98 & 81.97           & -       & -       \\
COAP~\cite{mihajlovic2022coap}      & 140            & 18.7 G       & 94.31 & 93.98 & 94.63     & - & - \\ \midrule %
VolumetricSMPL                                & \textbf{\phantom{0}15}   & \textbf{\phantom{1}3.1 G} & \textbf{94.67} & \textbf{94.25} & \textbf{95.10} & \textbf{\( 3.7 \times 10^{-5} \)} & \textbf{\( 3.5 \times 10^{-5} \)} \\ %
\bottomrule
\end{tabular}
}
\end{table}

\subsection{Volumetric Body Models} \label{subsec:exp_bodies} 
We train all models on mesh sequences from MoVi \cite{ghorbani2020movi} and DFaust \cite{dfaust:CVPR:2017} and evaluate them on the challenging PosePrior dataset \cite{PosePrior_Akhter:CVPR:2015}, which features extreme body poses, following the setup in \cref{subsec:training}. 
For evaluation, we report Mean Intersection over Union (IoU), Mean Squared Error (MSE) for SDF prediction, GPU memory usage, and inference speed.

\cref{tab:bodies} compares VolumetricSMPL with recent volumetric body models, LEAP~\cite{LEAP:CVPR:21} and COAP~\cite{mihajlovic2022coap}, both of which learn the SMPL shape space. Our model achieves higher accuracy while being 6× more memory efficient and around 10× faster in inference compared to COAP. This substantial reduction in computational cost makes VolumetricSMPL particularly advantageous for optimization and learning-based applications, where frequent volumetric body queries are required to impose penalty terms, as demonstrated in the next section. 

A direct comparison with imGHUM~\cite{alldieck2021imghum} is infeasible under our evaluation setup, as their released model is trained on private data and does not operate within the SMPL shape space. However, running their model shows that its inference is 86\% slower than VolumetricSMPL. 
For a fair comparison, in \cref{app_subsec:SMPL_prior}, we attempt to re-implement their architecture and train it under our setup. 
Notably, like LEAP, imGHUM lacks direct compatibility for resolving self-intersections (\cref{subsec:exp_selfinter}).

\begin{table}[tb]
  \caption{
    \textbf{Ablation Study} of design choices in our body model. 
    We compare the use of regular MLP instead of the proposed neural blend weights (NBW), the positional encoding ($\gamma(\cdot)$)~\cite{mildenhall2021nerf} (w. two levels) for the local query point, and the varying numbers of NBW components, from rank $R=1$ to $R=80$.
  }
  \label{tab:bodies_ablation}
  \centering
  \setlength{\tabcolsep}{4.0pt} %

\resizebox{\linewidth}{!}{%
\begin{tabular}{@{}lccc|ccc|cc@{}}
\toprule

& \multicolumn{3}{c}{$\text{Resources}$} 
& \multicolumn{3}{c}{$\text{IoU [\%]}~\uparrow$} 
& \multicolumn{2}{c}{$\text{MSE }{\scriptstyle[\times 10^{-5}]}~\downarrow$} \\
& $\text{t [ms]}~\downarrow$ 
& $\text{GPU}~\downarrow$ 
& $\text{\#param}$ 
& mean 
& \textit{surf.} 
& \textit{unif.} 
& SDF 
& $\lvert \text{SDF} \rvert$ \\
\midrule
Base MLP       & 15 & 2.9G & 0.4M & 92.75 & 92.23 & 93.27 & 5.2 & 6.9 \\
+ $\gamma(\cdot)$  & 15 & 3.1G & 0.4M & 93.00 & 92.42 & 93.59 & 8.3 & 11 \\ \midrule
+ NBW (R=1) & 15 & 3.1G & 0.8M & 94.06 & 93.55 & 94.58 & 4.8 & 5.2 \\
+ NBW (R=5) & 15 & 3.1G & 1.0M & 94.12 & 93.82 & 94.42 & 4.4 & 4.6 \\
+ NBW (R=10) & 15 & 3.1G & 1.2M &  94.07 & 93.67 & 94.46 & 4.1 & 3.9 \\
+ NBW (R=20) & 15 & 3.1G & 1.6M &  94.60 & 94.10 & 95.10 & 3.7 & 3.5 \\
+ NBW (R=40) & 15 & 3.1G & 2.4M &  \textbf{94.79} & 94.43 & 95.14 & 3.8 & 3.7 \\
+ NBW (R=80) & 15 & 3.1G & 4.0M &  94.67 & 94.25 & 95.10 & \textbf{3.7} & \textbf{3.5} \\

\bottomrule
\end{tabular}
}
\end{table}

We further analyze the impact of key design choices in \cref{tab:bodies_ablation}, evaluating:
\textit{1)} Our base method without Neural Blend Weights (NBW) (\cref{subsec:NBW}), instead using a standard MLP decoder. 
\textit{2)} The effect of positional encoding ($\gamma(\cdot)$) \cite{mildenhall2021nerf}, specifically retaining only low-frequency components (two levels).
\textit{3)} The number of NBW components $R$, ranging from 1 to 80. 

Following the evaluation setup in \cref{tab:bodies}, we observe that low-frequency $\gamma(\cdot)$ slightly improves IoU but reduces SDF accuracy, leading us to retain only two levels of $\gamma(\cdot)$. The most significant improvement comes from NBW, with performance saturating at $R = 80$, which we adopt as the final model configuration as it striks a good balance between IoU and SDF metrics. 
Notably, increasing the parameter count by increasing $R$ has a negligible impact on computational resources, as it does not expand the MLP decoder size while still improving learning capacity. 
Additional insights and ablation of model parameters are provided in \cref{app_subsec:ablation}. 

\subsection{Downstream Tasks} \label{subsec:exp_applications}

\subsubsection{In-the-Wild Human-Object Reconstruction}
\label{sec:hoi} 
Recovering 3D human-object layouts from a single image is crucial for action understanding. PHOSA~\cite{zhang2020phosa} introduced an optimization framework that refines human and object poses by enforcing contact rules. Wang \etal~\cite{wang2022reconstruction} improved scalability using large language models, but optimization remains costly, requiring $\sim$40 minutes (1k steps) due to repeated SDF computations for object meshes.

Instead of recomputing object SDFs at each optimization step, we represent the human as an SDF with VolumetricSMPL. This eliminates the bottleneck, accelerating optimization by $\sim$500× while maintaining or improving reconstruction accuracy.

Following \cite{wang2022reconstruction}, we evaluate collision (object penetration ratio) and contact (Chamfer distance) scores, along with per-iteration and total optimization time. As shown in \cref{tab:hoi}, VolumetricSMPL drastically reduces runtime ($\sim$500×) while preserving reconstruction quality. The test data is sourced from the “interaction database” introduced by \cite{wang2022reconstruction}, with metrics averaged over 100 images.

By improving efficiency and scalability, VolumetricSMPL paves the way for large-scale 3D human-object interaction dataset construction from web-scraped images.

\begin{table}[tb]
  \caption{\textbf{In-the-Wild Human-Object Reconstruction.} VolumetricSMPL achieves a 500× speedup in optimization time over \cite{wang2022reconstruction}, reducing per-iteration cost from 2128.2 ms to 4.28 ms and total optimization time from 35.9 min to 0.57 min. Despite this substantial acceleration, VolumetricSMPL maintains physical plausibility, achieving lower object penetration (collision) while preserving contact accuracy. We report optimization efficiency and reconstruction quality under identical settings, using the same optimization iterations and hardware (NVIDIA RTX 3090) on the ``interaction database’’ introduced by \cite{wang2022reconstruction}.
  }
\vspace{-0.5em}
\label{tab:hoi}
  \centering
  \setlength{\tabcolsep}{3.0pt} %

\resizebox{\linewidth}{!}{%

\begin{tabular}{@{}l|cc|cc@{}}
\toprule
& \multicolumn{2}{c}{Optimization Time $\downarrow$} & \multicolumn{2}{c}{Reconstruction Quality $\downarrow$} \\
 & Per Iteration & Total & Collision & Contact \\
\midrule
w. SDF \cite{wang2022reconstruction} & 2128.2 ms & 35.9 min & 6.33\% & \textbf{3.20} cm \\
w. Ours & \textbf{4.28} ms & \textbf{0.57} min & \textbf{5.73}\% & 3.96 cm \\
\bottomrule
\end{tabular}
}

\end{table}

\subsubsection{Human Mesh Reconstruction in 3D Scenes} \label{subsec:ego_hmr}
Reconstructing 3D human meshes from egocentric images is crucial for AR/VR applications.
Given a 3D scene and an egocentric image, the goal is to recover a physically plausible 3D human mesh.

Egocentric images often exhibit severe truncations due to close interaction distances, making pose inference challenging. To address this, we adopt the recent scene-conditioned diffusion model EgoHMR~\cite{zhang2023probabilistic}, which integrates COAP to guide sampling toward collision-free poses. However, COAP is computationally expensive, requiring substantial GPU memory and processing time, making inference costly, with over 2 seconds per frame.

By replacing COAP with VolumetricSMPL, we significantly improve efficiency while slightly enhancing accuracy. At each diffusion step, given a denoised SMPL body and a 3D scene point cloud, we compute signed distances via VolumetricSMPL to guide sampling toward collision-free configurations.

Following~\cite{zhang2023probabilistic}, we evaluate our method on the EgoBody dataset~\cite{zhang2022egobody}, measuring reconstruction quality for visible and occluded joints using MPJPE (Mean Per-Joint Position Error). Additionally, we assess physical plausibility with collision (\textit{coll.}) and contact (\textit{cont.}) scores, where collision measures scene penetration and contact evaluates proximity-based scene interaction. 

As shown in \cref{tab:egohmr-results}, replacing COAP with VolumetricSMPL enables EgoHMR to run with a 10× larger batch size on the same GPU (NVIDIA TITAN RTX, 24GB) while achieving over 3× faster inference per frame. Additionally, VolumetricSMPL reduces interpenetrations and provides a slight improvement in joint accuracy. These gains stem from VolumetricSMPL’s more efficient SDF-based gradients, which enable smoother and more precise collision handling compared to COAP’s occupancy step function.

\begin{table}[tb]
\centering
  \setlength{\tabcolsep}{3.0pt} %

\caption{
\textbf{Human Mesh Reconstruction in 3D Scenes} using EgoHMR with VolumetricSMPL vs. COAP. 
VolumetricSMPL enables 10× larger batch sizes (30 vs. 3) and 3.4× faster inference (0.61s vs. 2.08s per frame) while reducing human-scene collision (\textit{coll.}) (18.4\% vs. 19.1\%) and maintaining optimal contact (\textit{cont.}). MPJPE remains comparable for both visible (\textit{vis.}) and invisible (\textit{invis.}) joints, demonstrating improved efficiency without sacrificing reconstruction quality. 
Experiments conducted on an NVIDIA TITAN RTX (24GB) and the EgoBody dataset. 
}
\vspace{-0.5em}
\label{tab:egohmr-results}
\resizebox{\linewidth}{!}{%
\begin{tabular}{l|cc|cc|ccc}
\toprule
& \multicolumn{2}{c}{Resources} & \multicolumn{2}{c}{Interaction [\%]} & \multicolumn{3}{c}{MPJPE [mm] $\downarrow$} \\
& Batch $\uparrow$ & $t \downarrow$ & \textit{coll.}$\downarrow$ & \textit{cont.}$\uparrow$ & \textit{mean} & \textit{vis.}  & \textit{invis.} \\

\midrule
\textit{w.} COAP & 3 & 2.08 s & 0.191  & \textbf{99} & 86.74  & \textbf{65.7}  & 107.8 \\
\textit{w.} Ours & \textbf{30} & \textbf{0.61 s} & \textbf{0.184} & \textbf{99} & \textbf{86.63} & 65.8 & \textbf{107.5}  \\

\bottomrule[1pt]
\end{tabular}
}
\end{table}

\subsubsection{Scene-Constrained Human Motion Synthesis}
\label{subsec:human_motion}
Synthesizing realistic human motion in 3D scenes is crucial for applications in gaming, AR/VR, and robotics. A key challenge is ensuring the generated motion respects scene constraints, such as avoiding collisions, while preserving natural behavior.
The recent motion generation work DartControl~\cite{Zhao:DART:2024} introduces an optimization-based framework for synthesizing scene-constrained motions. It relies on evaluating human-scene collisions using precomputed scene SDFs to enforce collision constraints, which requires complete and high-quality scene scans. This limits its use to synthetic environments, as real-world reconstructions are often noisy or incomplete. Instead of modeling the scene as an SDF, we propose modeling the human as an SDF using VolumetricSMPL. This allows direct interaction with raw scans and enables optimization-based motion synthesis without requiring precomputed scene SDFs.

We evaluate this approach on an indoor navigation task, where the goal is to generate human motion from a start position to a goal while avoiding obstacles. We integrate VolumetricSMPL’s collision loss into DartControl’s optimization framework and compare it against a COAP-based baseline. Results, averaged over 15 reconstructed scenes (8 sequences each, 80–120 frames), are presented in \cref{tab:dart}. We evaluate:
\textit{1)}	Per-frame computation time and memory usage for collision evaluation.
\textit{2)}	Total resources required for optimizing an 80-frame sequence.
\textit{3)}	Motion quality, measured via scene collisions and final body-to-goal distance. 

VolumetricSMPL achieves a 7× speedup and 20× lower GPU memory usage per frame compared to COAP. 
Due to COAP’s high computational cost, scene collisions can only be evaluated for 16 frames per sequence on an 80GB GPU, requiring 71GB of memory. In contrast, VolumetricSMPL processes full 80-frame sequences on a 24GB GPU using only 15.9GB.

Notably, motion synthesized with VolumetricSMPL exhibits fewer scene penetrations and improved goal-reaching accuracy. This improvement is attributed to VolumetricSMPL’s SDF-based collision score, which provides smooth, informative gradients for optimization. In contrast, COAP’s occupancy-based collision score offers weaker guidance, often disrupting optimization and leading to suboptimal motion generation.

  \begin{table}[tb]
    \centering
      \setlength{\tabcolsep}{4.0pt} %
        
    \caption{
      \textbf{Scene-Constrained Human Motion Synthesis.} 
      Comparison of indoor navigation motion synthesis using~\cite{Zhao:DART:2024} with different collision evaluation methods. VolumetricSMPL significantly enhances motion quality and efficiency, reducing per-frame memory usage by 20× and computation time by 7× compared to COAP. For an 80-frame sequence, VolumetricSMPL enables full-sequence optimization on a single 24GB GPU, requiring only 15.9GB of memory. In contrast, COAP exceeds the capacity of an 80GB GPU, restricting collision checks to 16 frames (denoted with *) and still consuming 71GB. Motion quality is also improved, reducing scene collisions over 90\% and achieving more precise goal-reaching behavior. Experiments were conducted on an 80GB NVIDIA A100. 
    }

    \label{tab:dart}
    \resizebox{\linewidth}{!}{%
    \begin{tabular}{l|cc|c|cc}
    \toprule
    & \multicolumn{2}{c|}{Per-Frame $\downarrow$} & \multicolumn{1}{c|}{Total $\downarrow$} & \multicolumn{2}{c}{Motion Quality $\downarrow$} \\ %
    & Memory & Time & Memory & Collision & Goal Dist. \\
    \midrule
    \textit{w.} COAP & 4.44 GB & 26.53 ms & 71.0 GB* & 2.78 cm & 0.16 m \\
    \textit{w.} Ours & \textbf{0.19 GB} & \textbf{3.78 ms} & \textbf{15.9 GB} & \textbf{0.24 cm} & \textbf{0.01 m} \\
    \bottomrule[1pt]
    \end{tabular}
    }
  \end{table}

\subsubsection{Volumetric Constraints for Self-Intersections}
\label{subsec:exp_selfinter}
Self-intersections are a common issue in human mesh reconstruction. Like COAP, VolumetricSMPL leverages its compositional structure to impose volumetric constraints that resolve mesh intersections. Given a self-intersected body configuration ($\theta, \beta$), we optimize the pose $\theta^*$ to eliminate these intersections.

Following \cite{mihajlovic2022coap}, we approximate body parts using 3D bounding boxes $\mathcal{G}$ to efficiently detect potential collisions. Intersecting box pairs define overlapping volumes, from which we uniformly sample an initial set of points. This set is refined by selecting only points inside at least two body parts, determined via part-wise SDF predictions. Let this final set be $\mathcal{S}$; the self-intersection loss is then:
\begin{equation} \label{eq:self_pen_loss}
    \arg\min_{\theta^*} \sum\nolimits_{x \in \mathcal{S}} \sigma\left(f_\Theta(x|\theta,\beta)\right) + \mathcal{L}_{\text{PosePrior}},
\end{equation}
where $\mathcal{L}_{\text{PosePrior}}$ regularizes the optimization, ensuring the refined pose remains close to the original. Collisions between kinematically connected parts are ignored, as they are naturally expected to intersect (\eg, armpits).

We follow COAP’s experimental setup and evaluate our method on self-intersected human meshes sampled from PROX~\cite{PROX:2019}, using a subset of released 16 cases. The pose prior loss $\mathcal{L}_{\text{PosePrior}}$ is implemented as an $\ell_2$ loss between the initial and optimized pose. We retain COAP’s hyperparameters, setting the learning rate to $10^{-5}$ with a maximum of 200 optimization iterations, a pose prior weight of $10^3$, and a self-penetration loss weight of 0.1. The only change is replacing COAP’s volumetric self-penetration term with our implementation using VolumetricSMPL.

As shown in \cref{tab_app:abl_selfint}, VolumetricSMPL achieves per-step 2× faster optimization (14 ms \textit{vs.} 30 ms), 3× lower GPU memory usage (1.8 GB \textit{vs.} 6.2 GB), and overall fewer self-intersections (68 \textit{vs.} 74 triangle collisions~\cite{pavlakos2019expressive}). 
By significantly reducing computational overhead, VolumetricSMPL enables the efficient use of self-penetration constraints in downstream tasks that require larger batch sizes, making it more scalable for real-world applications.  
 
\begin{table}[t!]
\caption{
    \textbf{Volumetric Constraints for Self-Intersections.} 
    We compare VolumetricSMPL and COAP in terms of per-step computational cost (runtime and memory), total optimization iterations and time, and final mesh self-intersections. VolumetricSMPL achieves 2× faster per-step optimization (14ms vs. 30ms), 3× lower GPU memory usage (1.8GB vs. 6.2GB), and faster convergence (7.37 vs. 9.75 iterations, 0.10s vs. 0.29s total time). Additionally, it reduces self-intersections, producing fewer triangle collisions (68 vs. 74). These improvements demonstrate VolumetricSMPL’s superior efficiency and accuracy over COAP.
}

  \label{tab_app:abl_selfint}
  \centering
  \setlength{\tabcolsep}{5.0pt} %

\resizebox{\linewidth}{!}{%
\begin{tabular}{@{}l|cc|cc|c@{}}
\toprule
 & \multicolumn{2}{c}{Per-Step $\downarrow$} & \multicolumn{2}{c}{Total $\downarrow$} & \multicolumn{1}{c}{Mesh Triangle}\\ 
 & Time       & Memory  & Iterations & Time  & Collisions $\downarrow$ \\
\midrule
w. COAP             & 30 ms & 6.2 GB & 9.75 & 0.29 s & 74 \\ %
w. Ours   & \textbf{14 ms} & \textbf{1.8 GB} & \textbf{7.37} & \textbf{0.10 s} & \textbf{68} \\ %
\bottomrule
\end{tabular}
}

\end{table}

\section{Conclusion}
We introduced VolumetricSMPL, a novel volumetric body model that overcomes the computational limitations of prior neural implicit models while preserving accuracy and enabling efficient interaction modeling. Our approach leverages Neural Blend Weights Generator to significantly reduce the memory and compute requirements of SDF-based body representations, achieving 10× faster inference and 6× lower GPU memory usage compared to COAP.

Through extensive evaluations, we demonstrated that VolumetricSMPL improves both body model performance and downstream applications, including human mesh recovery, human-object interaction, scene-constrained motion synthesis, and self-intersection handling. By providing a scalable, differentiable, and computationally efficient framework for human interactions, VolumetricSMPL enables more realistic and robust modeling, paving the way for broader adoption in AR/VR, robotics, and autonomous systems. We hope this work serves as a foundation for future research in learning-based interaction modeling and advances in differentiable human-environment reasoning. 

\textbf{Acknowledgments.} 
Marko Mihajlovic is supported by Hasler Stiftung via the ETH Zurich Foundation (2025-FS-334).

{
    \small
    \bibliographystyle{ieee_fullname}
    \bibliography{egbib}
}

\clearpage
\begingroup

\twocolumn[
\begin{center}
    {\Large \bf \Large{\bf VolumetricSMPL:\\
    VolumetricSMPL: A Neural Volumetric Body Model for\\Efficient Interactions, Contacts, and Collisions} \\ 
    -- Supplementary Material -- \par}
  \vspace*{30pt}
\end{center}
]
\appendix
\setcounter{page}{1}
\setcounter{table}{0}
\setcounter{figure}{0}
\setcounter{equation}{0}
\counterwithin{figure}{section}
\counterwithin{table}{section}
\renewcommand{\thetable}{\thesection.\arabic{table}}
\renewcommand{\thefigure}{\thesection.\arabic{figure}}
\renewcommand{\theequation}{\thesection.\arabic{equation}}

\section{Overview} 
In this supplementary document, we provide additional implementation details (\cref{app_subsec:impl}) and further insights into the importance of design choices in VolumetricSMPL (\cref{app_subsec:ablation}). We also include additional information regarding the downstream applications of our model (\cref{sec:app:Downstream_tasks}).  

Next, we illustrate our easy-to-use Python interface, which seamlessly integrates with the widely used SMPL-X package (\cref{sec:app:code}). Finally, we discuss the limitations of VolumetricSMPL and outline potential future research directions (\cref{sec:app:limit_future}). 

\section{Implementation Details and Comparisons} \label{app_subsec:impl} 

\subsection{Network Architectures} 
The VolumetricSMPL architecture consists of two primary neural components: a shared PointNet~\cite{qi2017pointnet} encoder for all body parts and an MLP decoder, implemented using weights predicted by the Neural Blend Weights (NBW) generator (\cref{fig:NBW}). 

\textbf{VolumetricSMPL Encoder.} 
After a forward pass through the SMPL-based body model, the posed skin mesh is partitioned into 15 ($K=15$) body parts based on the kinematic chain of the human body, following~\cite{mihajlovic2022coap}. Each body part is then normalized according to its respective bone transformation  $G_k$. The local mesh of each part is resampled as a point cloud with 1k points and encoded using a shared PointNet encoder, ensuring memory-efficient alignment across all body parts. 

The shared encoder follows a 128-neuron MLP PointNet architecture, interleaved with ReLU activations. It consists of an input linear layer, four ResNet blocks, and an output layer. Each ResNet block contains two linear layers with a skip connection, facilitating effective feature propagation. The output layer produces a 128-dimensional latent code  $\mathbf{z}_k$, which conditions the decoder. This structure allows for efficient encoding of local shape variations while maintaining compact memory usage and fast inference speed. 

\textbf{VolumetricSMPL Decoder.} 
The MLP decoder is a compact 7-layer network with 64 neurons per layer and a skip connection on the 3rd layer, interleaved with ReLU activations. This lightweight architecture ensures efficient computation while maintaining high expressivity, as the Neural Blend Weights (NBW) framework enables a large number of learnable parameters to be utilized effectively.

In addition to being conditioned on the latent code, the MLP decoder also takes a local query point as input $\mathbf{x}_k$. To enhance spatial encoding, the query point is positionally encoded~\cite{mildenhall2021nerf} using only two frequency levels, providing better representation capacity for fine-scale details. Utilizing even higher frequency signals as input severely hampers the prediction accuracy and makes the training unstable. 

\subsection{Volumetric Bodies}
VolumetricSMPL is trained following the procedure outlined in \cref{subsec:training} of the main paper. For baseline comparisons, we use the publicly released COAP~\cite{mihajlovic2022coap} and LEAP~\cite{LEAP:CVPR:21} models, which are also trained on the AMASS subsets, for the respective SMPL~\cite{SMPL:2015,pavlakos2019expressive} body. 

\subsection{Additional Comparisons} \label{app_subsec:SMPL_prior} 
\begin{table}[t!]
\caption{
    \textbf{Comparison of Canonical SMPL-based vs. Flexible Direct Modeling.}  
    We evaluate the impact of conditioning VolumetricSMPL on an explicit mesh prior versus learning a volumetric representation directly in observation space using only low-dimensional shape and pose coefficients ($\beta, \theta$) as in \cite{alldieck2021imghum}.
    Both methods are trained under the same protocol (\cref{subsec:training}). Surf. and Unif. denote IoU scores computed for points sampled near the surface and uniformly in space, respectively.
    Results show that incorporating explicit mesh priors significantly improves IoU and reduces SDF prediction error, demonstrating the benefits of our canonical compositional modeling approach. The experimental setup follows \cref{tab:bodies}. 
}
  \label{tab_app:bodies}
  \centering
  \setlength{\tabcolsep}{5.0pt} %

\resizebox{\linewidth}{!}{%
\begin{tabular}{@{}l|ccc|cc@{}}
\toprule
& \multicolumn{3}{c}{$\text{IoU [\%]}~\uparrow$} 
& \multicolumn{2}{c}{$\text{MSE}~\downarrow$} \\
& mean 
& \textit{surf.} 
& \textit{unif.} 
& SDF 
& $\lvert \text{SDF} \rvert$ \\
\midrule
Direct Modeling \cite{alldieck2021imghum} & 39.64 & 32.28 & 47.01           & $2.5 \times 10^{-3}$       & $3.5 \times 10^{-4}$       \\
VolumetricSMPL & \textbf{94.67} & \textbf{94.25} & \textbf{95.10} & $\mathbf{3.7 \times 10^{-5}}$ & $\mathbf{3.5 \times 10^{-5}}$ \\ %
\bottomrule
\end{tabular}
}
\end{table}
 
Volumetric models such as VolumetricSMPL, LEAP \cite{LEAP:CVPR:21}, and COAP \cite{mihajlovic2022coap} are designed to share the same learning space as their underlying mesh-based parametric models \cite{SMPL:2015,pavlakos2019expressive}.
This ensures seamless integration with methods operating in the SMPL parameter space \cite{tian2023recovering}, as demonstrated in the applications section (\cref{subsec:exp_applications}).

In contrast, models such as NASA \cite{nasa} and imGHUM \cite{alldieck2021imghum} do not rely on explicit SMPL priors.
Instead, they learn volumetric representations directly from scans, body meshes, or a combination of both. While this increases flexibility, it tends to be computationally expensive—NASA requires per-subject training, while imGHUM relies on a significantly larger architecture and proprietary private human scans for training.

Specifically, imGHUM requires propagating a query point through a deep stack of MLPs: an 8-layer 512-dimensional MLP, an 8-layer 256-dimensional MLP, and two 4-layer 256-dimensional MLPs, being 86\% slower for inference compared to VolumetricSMPL. In contrast, our MLP decoder is significantly more lightweight, using a 7-layer 64-dimensional MLP for partitioned body parts.

\textbf{Impact of Excluding SMPL Priors.} 
To evaluate the role of explicit SMPL conditioning, we re-implement the imGHUM architecture under our training setup (\cref{subsec:training}).
This adaptation removes the PointNet encoder that processes SMPL-derived point samples, replacing it with a large SDF decoder MLP that directly conditions on SMPL parameters $(\beta,\theta)$.
This results in a volumetric model operating in observation space, without part-wise canonicalization.

We train both models and evaluate them following \cref{subsec:exp_bodies}.
Results in \cref{tab_app:bodies} show that excluding the explicit SMPL prior significantly degrades generalization when training data is limited.
Notably, our method requires training samples only within bounding boxes $\mathcal{B}$, leveraging an analytic SDF for the outer volume. 
Hence a model trained only with samples within $\mathcal{B}$ can produce floater artifacts in outside regions, leading to large errors.  

Direct comparison with the released pre-trained imGHUM model is infeasible, as it learns a different human shape space from proprietary data.
Additionally, an additional key advantage of compositional volumetric models (\eg, COAP and VolumetricSMPL) is their ability to resolve self-intersections (\cref{subsec:exp_selfinter}), unlike LEAP and imGHUM. 

\textbf{Performance Breakdown.} 
To better isolate the contributions of each component in our efficient querying pipeline (\cref{subsec:SDF_Querying}), we evaluated the impact of the coarse analytic SDF acceleration. In \cref{tab:bodies} and \cref{tab:bodies_ablation}, removing the coarse acceleration increases the runtime from 15\,ms to 25\,ms and memory usage from 3.1\,GB to 5.0\,GB—showing a $\sim$1.7$\times$ speedup and $\sim$1.6$\times$ memory reduction attributable to the analytic term.

\section{Ablation studies} \label{app_subsec:ablation}
\subsection{Impact of the Point Cloud Sampling} 
\label{app_subsec:ablation_pts} 
\begin{table}[t!]
\caption{
    \textbf{Impact of the Point Cloud Sampling.} 
    The number of point samples per-body part has moderate impact on the model performance and computational resources (inference speed and GPU memory). 1k samples is the default configuration in the main paper (denoted 1,000*). The models are trained for 10 epochs. The default setup strikes a good balance between accuracy and resource consumption. 
}
  \label{tab_app:abl_samples}
  \centering
  \setlength{\tabcolsep}{5.0pt} %

\resizebox{\linewidth}{!}{%
\begin{tabular}{@{}l|cc|ccc|cc@{}}
\toprule
Point 
& \multicolumn{2}{c}{$\text{Resources}$} 
& \multicolumn{3}{c}{$\text{IoU [\%]}~\uparrow$} 
& \multicolumn{2}{c}{$\text{MSE}~{\scriptstyle \times 10^{-5}}~\downarrow$} \\
Samples 
& $\text{t [ms]}~\downarrow$ 
& $\text{GPU}~\downarrow$ 
& mean 
& \textit{surf.} 
& \textit{unif.} 
& SDF 
& $\lvert \text{SDF} \rvert$ \\
\midrule
250     &15 &2.6 & 87.47 & 84.94 & 90.01 & 6.37 & 7.15\\ %
500     &15 &2.7 & 90.71 & 87.87 & 93.55 & 6.34 & 6.96\\ %
750     &15 &2.9 & 91.31 & 88.45 & 94.16 & 6.40 & 6.91\\ %
1,000*  &15 &3.1 & 91.43 & 88.49 & 94.38 & 6.47 & 6.96\\ %
1,250   &15 &3.2 & 91.45 & 88.51 & 94.38 & 6.52 & 6.97\\ %
1,500   &15 &3.4 & 91.40 & 88.50 & 94.30 & 6.56 & 7.00\\ %
1,750   &15 &3.5 & 91.33 & 88.44 & 94.21 & 6.57 & 7.00\\ %
2,000   &15 &3.7 & 91.33 & 88.42 & 94.24 & 6.59 & 7.06\\ %
\bottomrule
\end{tabular}
}
\end{table}

As described in the method section, VolumetricSMPL applies kinematic-based mesh partitioning~\cite{mihajlovic2022coap} to the SMPL body mesh and normalization of each mesh part according to its respective bone transformation  $G_k$. Each local mesh is then resampled into a point cloud with 1k points, which is subsequently encoded using a shared PointNet encoder to ensure efficient memory alignment across all body parts. 

To evaluate the impact of the number of sampled points on both model performance and computational efficiency, we conduct an ablation study, summarized in \cref{tab_app:abl_samples}. The reported results correspond to VolumetricSMPL trained for 10 epochs under the experimental setup outlined in the main paper. The findings indicate that the default setting of 1k points achieves a good balance between accuracy and resource efficiency, making it a suitable choice for practical deployment. 

\subsection{Impact of the Bounding Box Size} \label{app_subsec:ablation_boxsize} 
\begin{table}[t!]
\caption{
    \textbf{Impact of the Bounding Box Size.} 
    We evaluate how different bounding box sizes  $B_k$  affect SDF accuracy.
    Larger boxes degrade SDF predictions, as the neural network becomes less precise further from the iso-surface, making analytic SDF estimation preferable in these regions. 
    The optimal padding of 12.5\% (denoted by *) achieves the lowest mean SDF error and is used in the final model. Reported values are scaled by $\times 10^{-5}$. 
}
  \label{tab_app:abl_boxsize}
  \centering
  \setlength{\tabcolsep}{2.5pt} %
  \renewcommand{\arraystretch}{1.0} %

\resizebox{\linewidth}{!}{%

    \begin{tabular}{@{}l|cccccccccccccc@{}}
    \toprule

         & \multicolumn{14}{c}{$\text{Bounding Box } B_k \text{ Size (\%)}$} \\
        MSE                  & 5 & 7.5 & 10 & \textbf{12.5*} & 15 & 20 & 30 & 40 & 50 & 60 & 70 & 80 & 90 & 100 \\
        \midrule
        SDF               & 13.7 & 11.6 & 9.0  & 6.5  & 6.3  & 6.2  & 6.0  & 6.1  & 6.4  & 6.8  & 7.3  & 7.9  & 8.4  & 9.0  \\
        $\lvert \text{SDF} \rvert$ & 6.3  & 6.5  & 6.7  & 7.0  & 7.3  & 8.0  & 10.1 & 13.8 & 19.5 & 27.9 & 39.9 & 56.0 & 77.7 & 104.9 \\\hdashline
        \textit{mean}     & 10.0 & 9.1  & 7.9  & \textbf{6.7}  & 6.8  & 7.1  & 8.1  & 10.0 & 12.9 & 17.4 & 23.6 & 32.0 & 43.0 & 57.0  \\
        \bottomrule
    \end{tabular}
}
\end{table}

We evaluate how the bounding box size $B_k$ affects model performance. While varying the padding level does not meaningfully affect occupancy/sign evaluation, it does influence signed distance accuracy. 

To analyze this effect, we ablate padding levels from 5\% to 100\%, with results summarized in \cref{tab_app:abl_boxsize}.
As shown, larger bounding boxes degrade SDF predictions, as the neural network becomes less precise further from the iso-surface, making analytic SDF estimation preferable in these regions. 

The optimal padding of 12.5\% achieves the lowest mean SDF error which is adopted as the final model parameter. 

\subsection{Even smaller MLPs: Impact on Efficiency and Accuracy} 
\begin{table}[tb]
  \caption{
  \textbf{Impact of MLP Size on Efficiency and Accuracy.}  We evaluate the effect of reducing the MLP decoder size by comparing architectures with 32, 40, 50, and 64 neurons (default). As shown, smaller MLPs significantly reduce computational cost: using 32 neurons instead of 64 reduces inference time by 33.3\% (15 ms $\to$ to 10 ms) and GPU memory usage by 35.5\% (3.1 GB $\to$ 2.0 GB). However, this comes at the cost of increased SDF error ($|SDF|$ rising by $\sim$50\%) and decreased IoU. The 64-neuron configuration achieves a good balance between computational efficiency and reconstruction accuracy, making it the optimal choice for resource-intensive downstream tasks. 
  }
  \label{tab:app:mlp_size}
  \centering
  \setlength{\tabcolsep}{4.0pt} %
  \renewcommand{\arraystretch}{1.0} %

\resizebox{\linewidth}{!}{%
\begin{tabular}{@{}l|ccc|ccc|cc@{}}
\toprule

& \multicolumn{3}{c}{$\text{Resources}$} 
& \multicolumn{3}{c}{$\text{IoU [\%]}~\uparrow$} 
& \multicolumn{2}{c}{$\text{MSE }{\scriptstyle [\times 10^{-5}]}~\downarrow$} \\
Neurons 
& $\text{t [ms]}~\downarrow$ 
& $\text{GPU}~\downarrow$ 
& $\text{\#param}$ 
& mean 
& \textit{surf.} 
& \textit{unif.} 
& SDF 
& $\lvert \text{SDF} \rvert$ \\
\midrule

32 & 10 & 2.0G & 1.7M & 94.12 & 93.73 & 94.50 & 4.5 & 5.2 \\
40 & 11 & 2.2G & 2.1M & 94.02 & 93.60 & 94.45 & 4.6 & 5.2 \\ 
50 & 12 & 2.4G & 2.8M & 94.54 & 94.25 & 94.82 & 5.3 & 5.5 \\ \hdashline
64 & 15 & 3.1G & 4.0M & \textbf{94.67} & \textbf{94.25} & \textbf{95.10} & \textbf{3.7} & \textbf{3.5} \\
\bottomrule
\end{tabular}
}
\end{table}

We further analyze the impact of reducing the MLP decoder size by comparing architectures with 32, 40, and 50 neurons with the default setting of 64 neurons. Results are summarized in \cref{tab:app:mlp_size}. 

As shown in \cref{tab:app:mlp_size}, smaller MLPs reduce computational costs. Using 32 neurons instead of 64 reduces computation time by 33.3\% (15 ms $\to$ 10 ms) and GPU memory usage by 35.5\% (3.1 GB $\to$ 2.0 GB).  
However, this comes at the cost of substantially lower accuracy, with $\lvert \text{SDF} \rvert$ errors increasing by ~50\%.  
The 64-neuron setup achieves a good trade-off between computational efficiency and reconstruction accuracy while being useful for many resource intensive downstream tasks.

\subsection{Alternative Architectures}
We also experimented with alternative MLP architectures, including SIREN~\cite{sitzmann2020implicit}, HyperNetworks~\cite{Ha2017HyperNetworks}, and FiLM-based conditionings~\cite{chan2021pi}.  
However, due to the weak supervision signal in our model—where only the global SDF prediction (after the min operation) is supervised—these approaches failed to converge in a feed-forward setting.

\subsection{Comparison with Classic Techniques} 
We further compare our method against traditional techniques such as winding numbers~\cite{jacobson2013robust}, which have been used in human contact modeling~\cite{muller2024generative, fieraru2021remips}. However, winding numbers are not differentiable and do not generalize well to applications that require differentiable collision loss terms~\cite{zhang2023probabilistic, li2024interdance, wang2022reconstruction, zhang2020phosa}, unlike VolumetricSMPL.

To quantify the efficiency gap, we evaluate inference time and GPU memory usage for occupancy checks between two SMPL-X bodies by determining whether one body’s vertices reside inside another. We adopt the implementation from~\cite{muller2024generative} and report the results in \cref{tab:winding}.

As shown in \cref{tab:winding}, winding numbers introduce significant computational overhead, making them impractical for learning-based tasks requiring large batch sizes without down-sampling human meshes. In contrast, VolumetricSMPL is over 40× faster and consumes 50× less GPU memory, enabling efficient large-scale training and inference. 

\begin{table}[tb]
\centering
\setlength{\tabcolsep}{4.0pt} %
\caption{ \textbf{Comparison of occupancy checks using winding numbers and VolumetricSMPL.}
We evaluate inference time and GPU memory usage to check whether SMPL-X vertices are inside another SMPL-X body. VolumetricSMPL achieves over 40× faster inference and reduces GPU memory usage by 50×, making it significantly more efficient for large-scale learning tasks. } 
\label{tab:winding} 
\resizebox{\linewidth}{!}{
    \begin{tabular}{l|cc}
        \toprule
        & \multicolumn{2}{c}{Resources} \\ 
        & Inference Time $\downarrow$ & GPU Memory $\downarrow$ \\ 
        \midrule
        Winding Numbers \cite{jacobson2013robust,muller2024generative} & 464.41 ms & 15.58 GB \\ 
        VolumetricSMPL & \textbf{11.06} ms & \textbf{0.27} GB \\ 
        \bottomrule[1pt]
    \end{tabular}}
\end{table}

\begin{figure*}[t]
    \begin{center}
        \includegraphics[width=\textwidth]{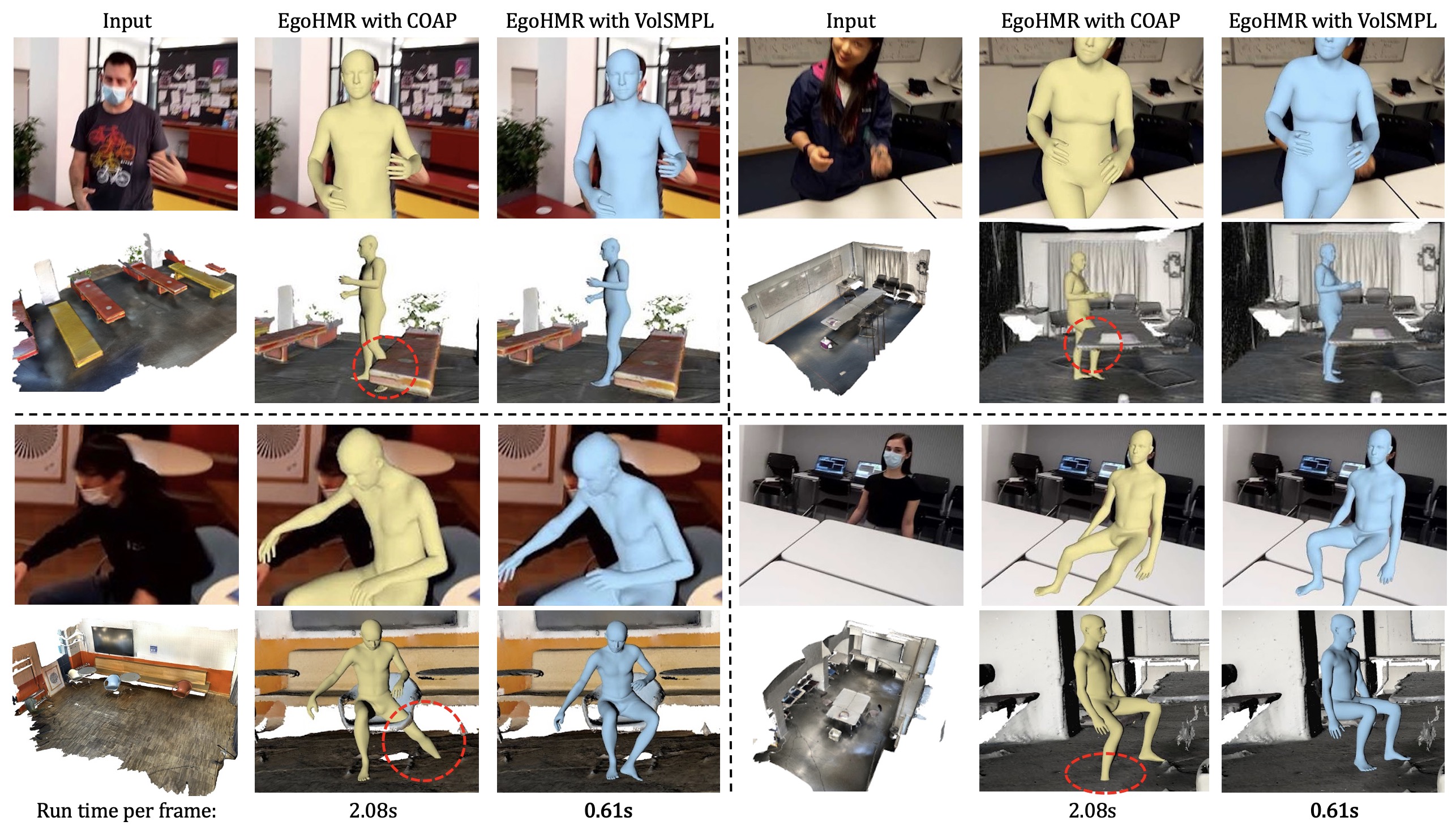}
    \end{center}
    \caption{\textbf{Human Mesh Recovery in 3D Scenes.} Given an egocentric image and the 3D scene mesh as input, EgoHMR with VolumetricSMPL (in blue) achieves fewer human-scene collisions than with COAP (in yellow) while being substantially faster (2.08s \textit{vs.} 0.61s). The collisions are denoted by the red circles.}
    \label{fig:egohmr}
\end{figure*}

\begin{figure*}[ht!]
    \begin{center}
        \includegraphics[width=0.98\textwidth]{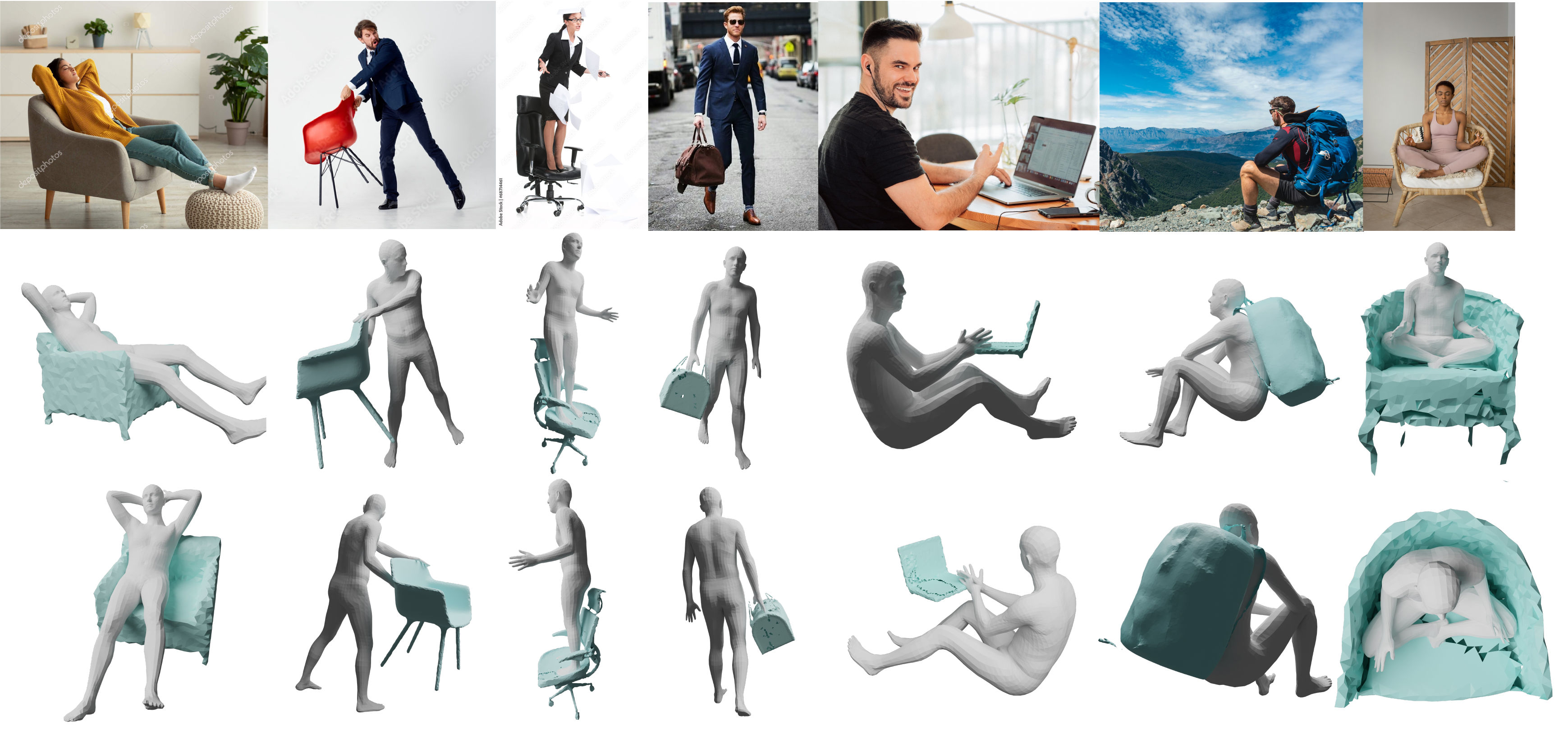}
    \end{center}
    
    \caption{\textbf{Reconstructing Human-Object Interactions from Images in the Wild.}    
    Here we demonstrate how VolumetricSMPL can be integrated to reconstruct human-object interactions from images in the wild~\cite{zhang2020phosa, wang2022reconstruction}. VolumetricSMPL achieves comparable reconstruction quality while being significantly faster than calculating human mesh SDF. See more details in \ref{sec:hoi}. 
    }
    \label{fig:hoi}
\end{figure*}

\section{Downstream Tasks}\label{sec:app:Downstream_tasks}
In the following, we provide further implementation details for the downstream tasks illustrated in \cref{fig:teaser}. 

\subsection{Reconstructing Human-Object Interactions from Images in the Wild} \label{sec:app:rhoi} 
Following the two-stage methodology proposed by Wang \etal\cite{wang2022reconstruction} and PHOSA~\cite{zhang2020phosa}, we first independently reconstruct humans and objects from the input image. In the second stage, a joint optimization step refines their contacts and spatial arrangements. 

Unlike~\cite{wang2022reconstruction}, which uses a time-consuming collision loss based on mesh-triangle intersections, we replace this step with an efficient alternative by transforming the body mesh into signed distance fields. Specifically, we use VolumetricSMPL to compute penetration loss efficiently, significantly improving computational speed while maintaining accuracy.

\textbf{Initial Body Poses and Shapes Estimation.} 
We use PARE~\cite{Kocabas_PARE_2021} to predict the pose  $\theta$  and shape  $\beta$  parameters of the SMPL~\cite{SMPL:2015} parametric body model. Next, we leverage MMPose~\cite{mmpose2020} to detect 2D body joint keypoints  $\mathbf{J}_{\mathit{2D}}$. Finally, we refine the predicted SMPL body model by fitting it to the detected 2D keypoints using SMPLify~\cite{Bogo:ECCV:2016}.

The optimization objective for SMPLify is formulated as:
\begin{equation}
\label{eq:smplify}
    E(\beta, \theta) = \Big|\Big| \Pi (\hat{\mathbf{J}}_{\mathit{3D}}) - \mathbf{J}_{\mathit{2D}} \Big|\Big|^{2}_{2} + E_{\theta} + E_{\beta},
\end{equation}
where $E_{\theta}$ and $E_\beta$ are pose and shape prior terms, $\hat{\mathbf{J}}_{\mathit{3D}}$ represents the estimated 3D body joints, and $\Pi$ is the perspective projection operator. 

\textbf{Initial Object Pose and Shape Estimation.} 
We formulate object pose and shape estimation as a rendered shape-matching problem. First, we select an object mesh from a set of template meshes for each object category, choosing the one that best matches the corresponding 2D image.

Next, we use PointRend~\cite{kirillov2019pointrend} to detect objects in the image, extracting their bounding boxes, segmentation masks, and semantic labels. Finally, we refine the 6-DoF object pose of the selected mesh using a differentiable renderer~\cite{kato2018renderer}. For further details, refer to PHOSA~\cite{zhang2020phosa}. 

\textbf{Human-Object Joint Optimization.} 
The joint optimization process refines human and object scales, translations, and rotations by minimizing the following objective function:
\begin{equation}
    \mathcal{L} = \lambda_1 \mathcal{L}_{contact} + \lambda_2 \mathcal{L}_{normal} \\ +
    \lambda_3 \mathcal{L}_{pen} 
    + \lambda_4 \mathcal{L}_{scale}, %
    \label{eq:optmization}
\end{equation}
where  $\mathcal{L}_{scale}$  penalizes deviations between the current human/object scale  $s^{\prime}_c$  and the prior scale  $\bar{s}_c$  obtained from large language models: 
\begin{equation}
    \mathcal{L}_{scale} = \left\Vert s^{\prime}_c - \bar{s}_c \right\Vert.
\end{equation}

The contact loss $\mathcal{L}_{contact}$  encourages plausible human-object interactions by minimizing the one-way Chamfer distance between contact pairs:
\begin{equation}
    \mathcal{L}_{contact} = \sum_{(i,j)\in\mathcal{I}}\mathds{1}(\mathbf{n}_{\mathcal{P}_{h}^i}, \mathbf{n}_{\mathcal{P}_{o}^j}) d_{\text{CD}}(\mathcal{P}_{h}^i, \mathcal{P}_{o}^j).
\end{equation}

The surface normal consistency loss  $\mathcal{L}_{normal}$  enforces alignment between interacting human and object surfaces:
\begin{equation}
        \mathcal{L}_{normal} = \sum_{(i,j)\in\mathcal{I}}\mathds{1}(\mathbf{n}_{\mathcal{P}_{h}^i}, \mathbf{n}_{\mathcal{P}_{o}^j})(1 + d_{\cos}(\mathbf{n}_{\mathcal{P}_{h}^i}, \mathbf{n}_{\mathcal{P}_{o}^j})),
\end{equation}
where  $d_{\cos}(\mathbf{a}, \mathbf{b})=\frac{\mathbf{a}\cdot \mathbf{b}}{|\mathbf{a}||\mathbf{b}|}$  is the cosine similarity between two normal vectors.

The penetration loss $\mathcal{L}_{pen}$ prevents object vertices from penetrating the body mesh:
\begin{equation}
    \mathcal{L}_{pen} = \frac{1}{|\mathcal{O}|} \sum\nolimits_{v \in \mathcal{O}} \textrm{ReLU}(-f^{\textrm{Vol.SMPL}}_{\Theta}(v / s'^h_c|\theta, \beta)),
\end{equation}
where $f^{\textrm{Vol.SMPL}}_{\Theta}$ represents the VolumetricSMPL body model, and  $\mathcal{O}$  denotes the object mesh. 
The object vertex  $v$  is rescaled with $\frac{1}{s'^h_c}$, where $s'^h_c$ is the current human scale. 
Since VolumetricSMPL predicts a scale-invariant signed distance field, we apply this rescaling when querying the signed distance. 

In our experiments, we set  $\lambda_{1,..,4} = [1e5, 1e3, 1e4, 1e3]$. When using SDF-based optimization~\cite{wang2022reconstruction}, we adjust  $\lambda_{3} = 1e3$  instead of  $1e4$ . The joint optimization runs for 1k steps. The number of object vertices varies from $1$k to $20$k, depending on the object category. We use the Adam optimizer with a learning rate of $2e-3$. 

Additional visual results are displayed in \cref{fig:hoi}

\subsection{Human Mesh Recovery in 3D Scenes} \label{sec:app:hmr}
Given an egocentric image  $\mathcal{I}$  containing a truncated human body and a corresponding 3D scene point cloud  $\mathbf{P} \in \mathbb{R}^{N \times 3}$  in the camera coordinate system, where  $N$ is the number of scene points, EgoHMR aims to model the conditional distribution of human body poses $p(\theta | \mathcal{I}, \mathbf{P})$. The goal is to generate body poses that naturally interact with the 3D scene while aligning with the image observations. The body translation $\boldsymbol{\gamma}$  and shape parameters  $\boldsymbol{\beta}$ are modeled deterministically. 
During diffusion inference, at each sampling step $t$, the denoiser $D$ predicts the clean body pose $\hat{\theta}_0$ from the sampled noisy pose $\theta_t$ at timestep $t$: 
\begin{equation}
    \hat{\theta}_0 = D(\theta_t, t, \mathcal{I}, \mathbf{P}).
\end{equation}
For further architecture details, refer to~\cite{zhang2023probabilistic}. The predicted pose  $\hat{\theta}_0$  is then noised back to  $\theta_{t-1}$  using the DDPM sampler~\cite{ho2020denoising}:
\begin{equation}
\label{eq:egohmr-sample}
   \theta_{t-1} \sim \mathcal{N}(\mu_t (\theta_t, \hat{\theta}_0) + a \Sigma_t \nabla \mathsf{J}(\theta_t), \Sigma_t),
\end{equation}
where $\mu_t (\theta_t, \hat{\theta}_0)$ is a linear combination of $\theta_t$ and $\hat{\theta}_0$, and $\Sigma_t$ is a scheduled Gaussian distribution~\cite{ho2020denoising}. The sampling process is guided by the gradient of a collision score  $\mathsf{J}(\theta)$, which mitigates human-scene interpenetrations. The guidance is modulated by  $\Sigma_t$  and a scale factor $a$. 

For EgoHMR with COAP (corresponding to the experiment setup w. COAP in \cref{tab:egohmr-results} of the main paper), the collision score is computed by checking whether each scene vertex is inside the human volume, using COAP~\cite{mihajlovic2022coap}:
\begin{equation}
\label{eq:coap_loss}
    \mathsf{J}(\theta) = \frac{1}{|\mathbf{P}|} \sum\nolimits_{q \in \mathbf{P}} \sigma(f^{\textrm{coap}}_{\Theta}(q|\mathcal{G})) \mathbb{I}_{f^{\textrm{coap}}_{\Theta}(q|\mathcal{G}) > 0},
\end{equation}
where $f^{\textrm{coap}}_{\Theta}$ stands for the COAP body model, and $\sigma(\cdot)$ stands for the sigmoid function.

For EgoHMR with VolumetricSMPL (corresponding to the experiment setup w. Ours in \cref{tab:egohmr-results} of the main paper), the collision score is computed using the signed distance field predicted by VolumetricSMPL for each scene vertex:
\begin{equation}
    \mathsf{J}(\theta) = \frac{1}{|\mathbf{P}|} \sum\nolimits_{q \in \mathbf{P}} \textrm{ReLU}(-f^{\textrm{VolumetricSMPL}}_{\Theta}(q|\mathcal{G})),
\end{equation}
where $f^{\textrm{VolumetricSMPL}}_{\Theta}$ denotes the proposed VolumetricSMPL body model.

\textbf{Experiment Details.} 
In \cref{eq:egohmr-sample}, we set the scale factor $a$ to $0.4$ for EgoHMR with COAP and $30$ for EgoHMR with VolumetricSMPL. 
The diffusion sampling process consists of $50$ steps, with collision score guidance applied only during the last $10$ steps. In the final $5$ denoising steps, we scale  $\nabla \mathsf{J}(\theta_t)$ by $a$ only, omitting  $\Sigma_t$  to prevent the collision guidance from diminishing too early in the process. 

To ensure a fair comparison between COAP and VolumetricSMPL, we compute collision scores in \cref{eq:egohmr-sample} using 20k scene vertices sampled within a 2×2m cube centered around the human body. 

For evaluation, we use the official checkpoint from~\cite{zhang2023probabilistic} to perform diffusion sampling and evaluate on the EgoBody~\cite{zhang2022egobody} test set, which consists of 62,140 frames. For the further details about the evaluation metrics we refer the reader to \cite{zhang2023probabilistic}. 

Additional visual results are displayed in \cref{fig:egohmr}.

\subsection{Scene-Constrained Human Motion Synthesis} \label{sec:app:motion_app}

We use DartControl~\cite{Zhao:DART:2024} to generate scene-constrained navigation motion in 15 scanned scenes from the Egobody~\cite{zhang2022egobody} dataset.
Given the starting location and goal location in a 3D scene, we initialize the human with a standing pose and use the optimization-based motion synthesis method of DartControl to drive the human to reach the goal location while avoiding scene obstacles.
The 3D scenes are represented as point clouds for collision evaluation, with each point cloud containing 16384 points sampled from the original scan using farthest point sampling.
The motion sequences vary in length, ranging from 80 to 120 frames. We condition the locomotion style of all sequences using the text prompt ``walk".
The optimization objective for scene-constrained motion synthesis encourages the body pelvis of the last frame to reach the goal location and penalizes all detected human-scene collisions as follows:

\begin{equation}
    \mathcal{L} = \mathcal{F}(\mathbf{p}, \mathbf{g}) + w * \mathcal{L}_{coll},
\end{equation}
where $\mathbf{p}$ denotes the last frame body pelvis, $\mathbf{g}$ denotes the goal location, $\mathcal{F}$ denotes the smooth L1 loss \cite{girshickFastRCNN2015}, $w$ is a tunable weight for collisions, and $\mathcal{L}_{coll}$ is the scene collision term that we separately implement using COAP and VolumetricSMPL following prior task (\cref{subsec:ego_hmr}). 

We use a collision weight of $w=1$ for the VolumetricSMPL collision term and conduct experiments with varying collision weights for the COAP baseline. Our observations reveal that the COAP baseline struggles to effectively balance accurate goal-reaching and collision avoidance, leading to performance that is inferior to VolumetricSMPL, as demonstrated in Tab.~\ref{tab:coll_weights}. Notably, applying a large collision weight $w=1$ for COAP disrupts the optimization process, leading to a failure to resolve collisions and causing deviations from the intended goal location.

  \begin{table}[tb]
    \centering
      \setlength{\tabcolsep}{4.0pt} %
        
    \caption{
      Comparison of indoor navigation motion synthesis using COAP-based collision term with different weights and VolumetricSMPL-based collision term. 
    }

    \label{tab:coll_weights}
    \resizebox{\linewidth}{!}{%
    \begin{tabular}{l|cc|cc}
    \toprule
    & \multicolumn{2}{c|}{Per-Frame $\downarrow$} & \multicolumn{2}{c}{Motion Quality $\downarrow$} \\ %
    & Memory & Time & Collision & Goal Dist. \\
    \midrule
    \textit{w.} COAP ($w=0.01$) & 4.44 GB & 26.48 ms  & 4.92 cm & 0.02 m \\
    \textit{w.} COAP ($w=0.1$) & 4.44 GB & 26.53 ms  & 2.78 cm & 0.16 m \\
    \textit{w.} COAP ($w=1.0$) & 4.44 GB & 26.77 ms  & 4.92 cm & 0.57 m \\\hdashline
    \textit{w.} Ours ($w=1.0$) & \textbf{0.19 GB} & \textbf{3.78 ms}  & \textbf{0.24 cm} & \textbf{0.01 m} \\
    \bottomrule[1pt]
    \end{tabular}
    }
  \end{table}

\subsection{Self-Intersection Handling with VolumetricSMPL via Volumetric Constraints} 
When resolving self-intersections using volumetric constraints (\cref{subsec:exp_selfinter}), the model first detects potential collisions by enclosing each body part within a 3D bounding box $\mathcal{G}$. For every pair of intersecting boxes, the overlapping volume is identified, and 300 points are uniformly sampled within this region. These points are further refined by retaining only those that reside inside at least two body parts, as determined through part-wise SDF evaluations. The final set of valid intersection points is denoted as $\mathcal{S}$, and the self-intersection loss is computed according to \cref{eq:self_pen_loss}. 

In this experiment, we minimize the final loss term (Eq.~\ref{eq:self_pen_loss}) using the SGD optimizer to iteratively refine the pose parameters and resolve intersections effectively. The computational resources reported in \cref{tab_app:abl_selfint} are estimated on an NVIDIA RTX 3090 GPU card.

\section{Limitations and Future Work} \label{sec:app:limit_future} 
While VolumetricSMPL achieves a 10× speedup and 6× lower memory usage compared to prior work~\cite{mihajlovic2022coap}, further optimizations remain an important direction. Currently, it supports batch sizes up to 80 for human-scene interaction tasks (\cref{subsec:human_motion}) on a 24GB GPU, but this remains a bottleneck when modeling longer human motion sequences. Future work could explore memory-efficient architectures to further scale motion synthesis. 

Additionally, similar to other volumetric body models~\cite{LEAP:CVPR:21,mihajlovic2022coap}, VolumetricSMPL does not explicitly model detailed hand articulation, primarily due to limitations in available training data. A potential extension involves developing a specialized volumetric hand model and integrating it into our framework, enabling more precise hand-object interactions, particularly in fine-grained manipulation tasks.  

Beyond human modeling, our NBW formulation is inherently generic and can be applied to non-human shapes. Exploring its potential for learning articulated animal models, robotic structures, or generic deformable objects could extend its applicability beyond human-centric tasks. We leave this exploration for future work.  

By addressing these limitations, VolumetricSMPL could further improve efficiency, extend its scope to finer interactions, and generalize beyond human body modeling to broader applications in graphics, robotics, and virtual environments. 

\textbf{Broader Impact.}  
Beyond the immediate applications explored in this work, VolumetricSMPL has the potential to serve as a valuable tool for the broader research community. Its efficiency, scalability, and ease of integration make it suitable for a wide range of interaction tasks. By providing an open-source implementation, we aim to facilitate further research into volumetric representations, encourage new applications in dynamic human-scene interactions, and inspire extensions to non-human shapes. 

We hope that VolumetricSMPL will enable researchers and practitioners to advance human body modeling research and its downstream applications. 

\section{Seamless SMPL Code Integration} \label{sec:app:code} 
VolumetricSMPL is a lightweight and user-friendly add-on module for SMPL-based body models, enabling seamless volumetric extension. 

With just a single line of code, users can extend SMPL models with volumetric functionalities. After completing the forward pass, they gain access to key volumetric functionalities, including SDF queries, self-intersection loss, and collision penalties.
This implementation maintains full compatibility with existing SMPL-based reconstruction and perception applications. 

The following code snippet demonstrates how to install VolumetricSMPL and integrate it with an SMPL model to utilize its volumetric functionalities: 
\begin{lstlisting}[caption={VolumetricSMPL Installation via PyPi}, label={lst:volsmpl_install}]
pip install VolumetricSMPL
\end{lstlisting}
\begin{lstlisting}[caption={Integrating VolumetricSMPL with SMPL.}, label={lst:volsmpl}]
import smplx
from VolumetricSMPL import attach_volume

# Create an SMPL body
model = smplx.create(**smpl_parameters) 

attach_volume(model) # extend with VolumetricSMPL

# SMPL forward pass
smpl_output = model(**smpl_data)  

# Access volumetric functionalities
# 1) Query SDF for given points
model.volume.query(smpl_output, scan_points)

# 2) Compute self-intersection loss
model.volume.selfpen_loss(smpl_output)

# 3) Compute collision loss
model.volume.collision_loss(smpl_output, points)
\end{lstlisting}

The \texttt{attach\_volume()} function seamlessly extends any \emph{SMPL, SMPL-H, or SMPL-X} model with volumetric capabilities. Once the full forward pass is completed, users can efficiently compute signed distance field (SDF) queries, self-penetration loss, and collision penalties for physically plausible human interactions.

VolumetricSMPL is released under the MIT license and will be publicly available to the research community.

\clearpage

\end{document}